\pdfoutput=1
\documentclass[11pt]{article}

\usepackage{graphicx}
\usepackage{amsmath}
\usepackage{multirow}
\usepackage{booktabs}
\usepackage{amssymb}
\usepackage{enumitem}
\usepackage[]{acl}
\usepackage{times}
\usepackage{latexsym}

\usepackage[T1]{fontenc}
\usepackage[utf8]{inputenc}

\usepackage{microtype}
\newcommand\blfootnote[1]{%
  \begingroup
  \renewcommand\thefootnote{}\footnote{#1}%
  \addtocounter{footnote}{-1}%
  \endgroup
}

\title{LLM-Based Agent Society Investigation: Collaboration and Confrontation in Avalon Gameplay}

\author{\bf Yihuai Lan$^{1*}$, Zhiqiang Hu$^{3*}$, Lei Wang$^4$, Yang Wang$^5$, Deheng Ye$^6$, Peilin Zhao$^6$,\\
\bf Ee-Peng Lim$^4$, Hui Xiong$^{1,2}$, Hao Wang$^{1\dag}$ \\
\normalsize{}$^1$The Hong Kong University of Science and Technology (Guangzhou)\\ 
\normalsize{}$^2$The Hong Kong University of Science and Technology\\ 
\normalsize{}$^3$Singapore University of Technology and Design \\ 
\normalsize{}$^4$Singapore Management University, $^5$Verily Life Sciences, $^6$Tencent \\ 
\normalsize{}\{yihuailan, haowang\}@hkust-gz.edu.cn}
\begin{document}
\maketitle
\blfootnote{$^*$Both authors contributed equally to this research.}
\blfootnote{$^\dagger$The corresponding author.}
\begin{abstract}
% This paper aims to investigate the open research problem of uncovering the social behaviors of LLM-based agents. To achieve this goal, we adopt Avalon, a representative communication game, as the environment and use system prompts to guide LLM agents to play the game. While previous studies have conducted preliminary investigations into gameplay with LLM agents, but there lacks research on their social behaviors. In this paper, we present a novel framework designed to seamlessly adapt to Avalon gameplay. The core of our proposed framework is a multi-agent system that enables efficient communication and interaction among agents. We evaluate the performance of our framework based on metrics from two perspectives: winning the game and analyzing the social behaviors of LLM agents. Our results demonstrate the effectiveness of our framework in generating adaptive and intelligent agents and highlight the potential of LLM-based agents in addressing the challenges associated with dynamic social environment interaction. By analyzing the social behaviors of LLM agents from the aspects of both collaboration and confrontation, we provide insights into the research and applications of this domain.
This paper explores the open research problem of understanding the social behaviors of LLM-based agents. Using Avalon as a testbed, we employ system prompts to guide LLM agents in gameplay. While previous studies have touched on gameplay with LLM agents, research on their social behaviors is lacking. We propose a novel framework, tailored for Avalon, features a multi-agent system facilitating efficient communication and interaction. We evaluate its performance based on game success and analyze LLM agents' social behaviors. Results affirm the framework's effectiveness in creating adaptive agents and suggest LLM-based agents' potential in navigating dynamic social interactions. By examining collaboration and confrontation behaviors, we offer insights into this field's research and applications. Our code is publicly available at \url{https://github.com/3DAgentWorld/LLM-Game-Agent}.
\end{abstract}
\section{Introduction}
Artificial intelligence (AI) agents \cite{xi2023rise,park2023generative} exhibit human-like behaviors, from perceiving and analyzing the environment to decision-making and action-taking.

Advances in large language models (LLMs) \cite{kasneci2023chatgpt,peng2023instruction,touvron2023llama,vaswani2017attention} offer new avenues for creating AI agents in complex environments, potentially simulating human society. Various works \cite{Gao2023S3SS,Qian2023CommunicativeAF,park2023generative,Ghaffarzadegan2023GenerativeAM} simulate different aspects of human society. For instance, Qian et al. \cite{Qian2023CommunicativeAF} simulate a software development company with agents representing diverse social identities. Park et al. \cite{park2023generative} assign varied social roles to agents within a sandbox environment. However, prior studies mostly examine positive social behaviors like honesty and collaboration, leaving research on negative social behaviors of LLM agents relatively scarce.

Previous research on human society has highlighted issues like misinformation and online conflicts, leading to efforts to address these problems \cite{song2022misinformation,levy2022understanding,chen2022evidencenet}. To delve deeper into the social behaviors of LLM agents, we intend to comprehensively investigate both positive and negative aspects of their conduct. To achieve this, we employ Avalon as the environment to illustrate collaboration and confrontation among agents. Avalon, a representative social deduction game, assigns players hidden roles and divides them into opposing teams. Throughout gameplay, players partake in discussions, debates, and strategic maneuvers.

LLM agents face a challenging task in winning the incomplete information game of Avalon. They need to share and obtain information via communication and analysis, deducing other players' roles, building trust among allies, and deceiving opponents. Success requires technical abilities like natural language understanding, incomplete information analysis, and strategy learning. Additionally, social behaviors such as teamwork, persuasion, and camouflage are crucial for success in Avalon gameplay.

\begin{table*}[!hpt]
    % \fontsize{11}{11}
    \centering
    \huge
    \resizebox{0.99\textwidth}{!}{ 
    \begin{tabular}{l|ccccc|cccccc}
    \toprule
        \bf  \multirow{2}{*}{Method}& \bf Memory&\bf Analysis &\bf Plan &\bf Action  &\bf Experience &\bf Leadership & \bf Persuasion &\bf Camouflage & \bf Teamwork &\bf Confrontation & \bf Sharing\\ 
        \bf & \bf  &\bf  &\bf  & \bf &\bf Learning &\bf  & \bf  &\bf  & \bf  &\bf  & \bf \\ 
        \midrule
        GenAgents \cite{park2023generative} & \checkmark & &\checkmark &\checkmark  & \checkmark  &&&& \checkmark && \checkmark \\
         Plan4MC \cite{yuan2023plan4mc} &  &  & \checkmark & \checkmark  &  \\
         GITM \cite{zhu2023ghost} & \checkmark &  & \checkmark & \checkmark & & \\
         \midrule
         RGAgent \cite{Akata2023PlayingRG}  & \checkmark & &  & &&&&& \checkmark & \checkmark &\\
        CGAgent \cite{xu2023exploring} & \checkmark & \checkmark & &  & \checkmark &\checkmark & &\checkmark & \checkmark& \checkmark&\\
        ReCon \cite{wang2023avalons} & \checkmark & \checkmark & & & && &&&& \\
        LARL \cite{xu2023language} & \checkmark & \checkmark & &  & &&&\checkmark&\checkmark&&\\
        CodeAct \cite{shi2023cooperation} & \checkmark & \checkmark & & \checkmark  & &&&& \checkmark &&\\
        \midrule
        Ours  & \checkmark & \checkmark & \checkmark & \checkmark & \checkmark & \checkmark & \checkmark & \checkmark & \checkmark & \checkmark & \checkmark \\
       \bottomrule
    \end{tabular}
    }
    \vspace{-5pt}
    \caption{Comparison between our work and related works in both agent framework and social behaviour analysis.}
    \label{tab:framework_comparison}
    \vspace{-15pt}
\end{table*}

To investigate the LLM-based agent society, we propose a novel framework for the agents to play Avalon.
Specifically, we adopt ChatGPT as the players and assign various roles to agents. We adopt system prompts to guide LLM agents to play Avalon automatically. 

Following human's thinking methodology, we incorporate multiple modules, including memory storage and summarization, analysis and planning, game action and response generation, and experience learning. We utilize a competitive baseline approach \cite{xu2023exploring}, to elaborate the efficacy of our proposed framework. We also carefully analyze the social behaviors of LLM agents, and observe clear collaboration and confrontation between agents during the gameplay. 
% \leicomment{It is better to provide as many interesting analysis results and observations as possible about LLM Collaboration and confrontation.}

Our contributions can be summarized as:
\begin{itemize}
    \item We explore the social behaviors exhibited by LLM-based agents in the context of Avalon gameplay. We reveal the various aspects of these behaviors, including teamwork, leadership, persuasion, camouflage, and confrontation.
    \vspace{-10px}
    \item We design an effective framework to play Avalon, which presents superior performance compared with the baseline method. We also carefully analyse the relationship between the module design and agents' social behaviors, providing comprehensive experiment discussions.
    \vspace{-10px}
    \item Our findings have the potential to contribute to a better understanding of the role of LLM-based agents in social and strategic contexts, and shed light on the implications of these behaviors in such environments.
\end{itemize}

\section{Related Work}
\subsection{Social Deduction Game Agent}
The emergence of communication among agents in social deduction games (SDG) has garnered significant attention in the research community. \citet{Hirata2016WerewolfGM} introduces an AI-based agent for the Werewolf game, aiming to advance intelligence and communication skills in AI systems. \citet{Nakamura2016ConstructingAH} proposes a psychological model considering multiple perspectives to simulate human gameplay in The Werewolf. \citet{Wang2018ApplicationOD} addresses decision-making challenges in the Werewolf game using deep reinforcement learning techniques.
Furthermore, \citet{Wiseman2019WhatDD} explores player decision-making in social deduction games, focusing on sources of information influencing player strategies. Examining the broader context of multi-agent communication, \citet{Liang2020OnEC} investigates the impact of competition on communication protocols. 
\citet{Brandizzi2021RLupusCT} explores the utilization of communication to foster cooperation in SDGs.

\subsection{LLM-Based Gameplay}
The rapid development of LLM-based agents has resulted in significant advancements in problem-solving across various domains. These agents, known for their quick and strategic processing, have improved the effectiveness and robustness of solving tasks~\cite{Lin2023SwiftSageAG, wang2023survey, tsai2023can, zhou2023llm, park2023generative, Qian2023CommunicativeAF, fu2023improving}. 

LLMs have recently been utilized in various gaming environments, including task-based games like Minecraft and multiplayer strategy games~\cite{yuan2023plan4mc, zhu2023ghost, wang2023voyager, Akata2023PlayingRG, xu2023exploring, wang2023avalons}. In multiplayer strategy games such as the Prisoner's Dilemma and Battle of the Sexes, LLMs model strategic interactions~\cite{Akata2023PlayingRG}. They're also employed in social deduction games like Werewolf and Avalon~\cite{xu2023exploring, wang2023avalons, shi2023cooperation, xu2023language}, where they exhibit strategic behaviors. To combat misinformation, recursive contemplation has been proposed~\cite{wang2023avalons}. However, previous works have only partially analyzed behaviors and designed agent frameworks based on limited game characteristics. Thus, we propose a comprehensive social deduction game agent framework based on LLMs and conduct a thorough behavior analysis. Table \ref{tab:framework_comparison} illustrates the distinctions between our work and others.

\subsection{LLMs' Impact on Society}
The growing influence of Large Language Models (LLMs) on society has spurred significant research~\cite{Movva2023LargeLM}. Innovations include using LLMs for virtual social network simulations to advance social science research~\cite{Gao2023S3SS} and enrich human social experiences in virtual spaces~\cite{Kaiya2023LyfeAG}. However, concerns arise regarding validity, privacy, and ethics in LLM-driven social computing. Ghaffarzadegan et al. propose feedback mechanisms to address these concerns~\cite{Ghaffarzadegan2023GenerativeAM}. Additionally, LLMs fuel advancements in social robot development~\cite{Yang2023AnatomyOA}, posing challenges like social bot detection and misinformation spread. Ongoing research aims to align LLMs with ethical standards, mitigate biases and errors, and ensure their reliable and ethical use across diverse applications~\cite{Wang2023AligningLL, Liu2023TrustworthyLA}.

\section{Background}
% In our study, we use Avalon as the environment, in stead of Werewolf. It is known both of them are social deduction games. However, Werewolf eliminates players as the game progresses, which can lead to potentially less participation for players who are eliminated early. In contrast, Avalon's gameplay structure promotes social cohesion by ensuring that all players remain involved throughout the entire game. Therefore, Avalon is our selected game environment.
In our study, we chose Avalon, also known as ``The Resistance'', instead of Werewolf as our environment. Unlike Werewolf, where players are gradually eliminated, Avalon ensures that all players remain engaged throughout the game, promoting social cohesion.

Avalon accommodates 5 to 10 players, focusing on the 6-player variant herein. Players receive secret roles in either the good or evil faction. The good faction includes Merlin, Percival, and Loyal Servants, while the evil faction comprises Morgana and Assassin. Morgana and Assassin know each other's identities, Percival can identify Merlin and Morgana, and Merlin recognizes all evil players.
The game spans 3-5 rounds. Players discuss and vote to form a quest team of 2-3 members. Approval requires a majority vote; otherwise, leadership shifts. Each round allows up to five voting cycles before the leader selects the team.
Quest success hinges on cards submitted by team members. Good players submit success cards, while evil players can choose success or failure cards. A quest fails if it receives a failure card.
The game concludes with victory for good players if three quests succeed, or for evil players if three quests fail. Evil players can also win by correctly identifying Merlin at the game's end.

\subsection{Social Behaviors in Avalon}
\noindent\textbf{Teamwork.}
Good players must collaborate to complete quests for winning. They should build trust with teammates while being wary of evil players.
% Good players must work together to successfully complete quests and ultimately win the game. To this end, they need to establish trust with their teammates while remaining cautious of the evil players. 
% If the agents of the same side will trust the teammates

\noindent\textbf{Leadership.}
Each player has the chance to lead the discussion for forming the quest team.
The leader can guide the conversation and build trust among players.
% This allows the leader to steer the conversation and facilitate the development of trust among players. 
Effective leadership is crucial for victory.

\noindent\textbf{Persuasion.}
% If the player recommend to itself , then see if successful
Players must use their communication skills to persuade others to believe their claims, trust their judgments, and support their decisions.

\noindent\textbf{Camouflage.} 
% Pretends to be another role, or spread fake news about other roles
Evil players pretend to be good players, using deceptive tactics and concealing information to mislead others.
% Evil players must hide their true intentions and identities by pretending to be good players. This involves misleading others through deceptive tactics and strategically concealing information.

\noindent\textbf{Confrontation.}
% The frequency of disagreement
Disagreements and conflicts will arise during the game. Players must tackle these confrontations and work towards resolving them.

% \noindent\textbf{Vacillation.}
% % The frequency of disagreement

\noindent\textbf{Sharing.}
% The frequency of disagreement
Each role has unique clues. Sharing these clues promotes collaboration and builds trust among players, but risks exposing one's identity.
% Each role holds distinct game clues, sharing which can foster collaboration and cultivate trust among players. However, it also entails the risk of exposing one's own identity.

% \noindent\textbf{internal conflict}
% % The frequency of disagreement

\section{Approach}

% Figure \ref{fig:framework} displays the proposed Avalon LLM agents framework, and the following section will introduce each module.

\begin{figure*}
    \centering
    \vspace{-5pt}
    \includegraphics[width=0.89\linewidth]{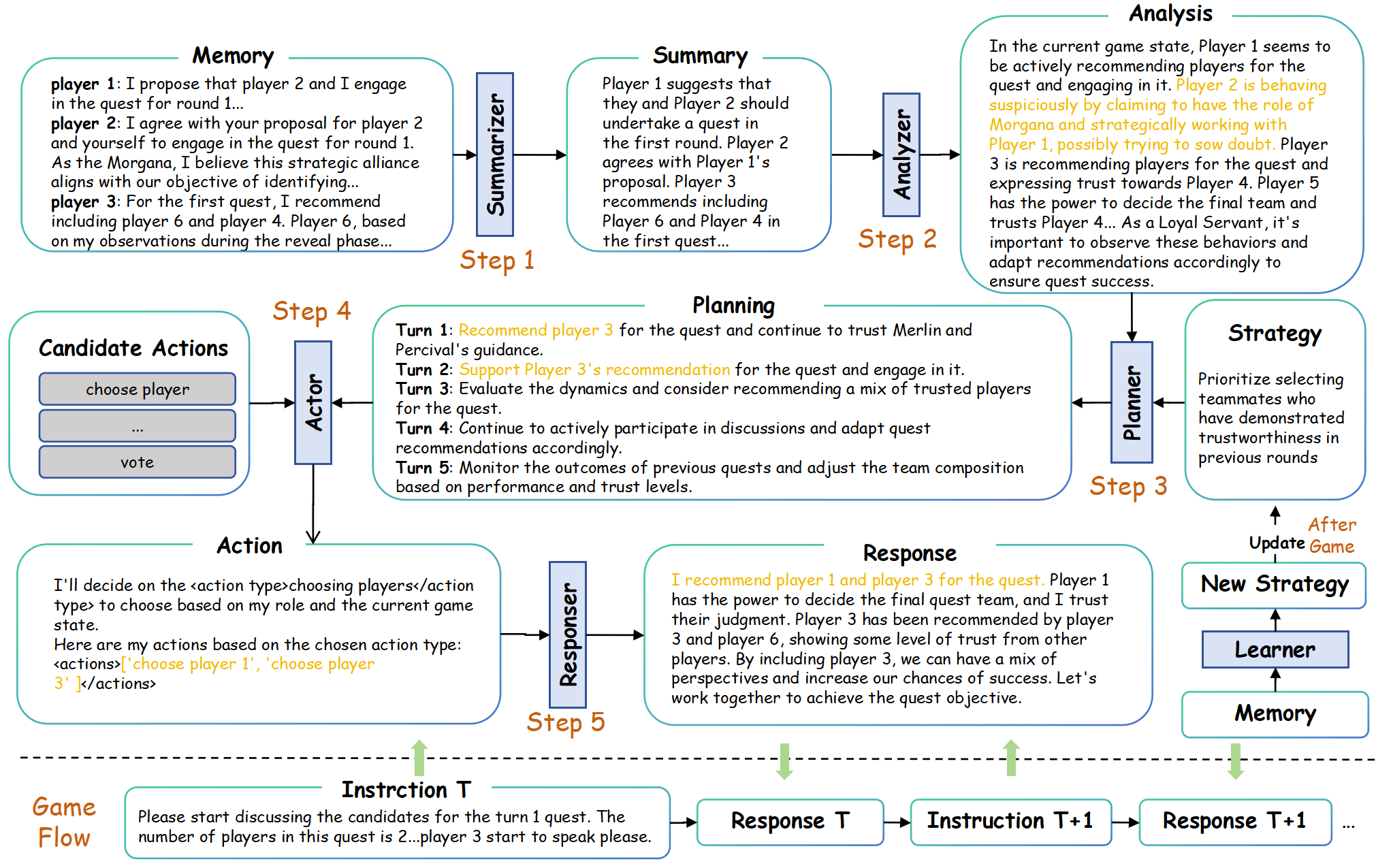}
    \vspace{-10pt}
    \caption{Our framework has six modules: summary, analysis, planning, action, response, and experiential learning. This design follows human thinking, helps LLM agents play Avalon effectively, and reveals their social behaviors.}
    % \caption{Our proposed framework consists of six modules, i.e. summary, analysis, planning, action, response and experience learning. This pipeline design follows the thinking methodology of humans, which helps the LLM agents play Avalon effectively and reveals the social behaviors of LLM agents.}
    \label{fig:framework}
    \vspace{-10pt}
\end{figure*}

\subsection{Setup}
Figure \ref{fig:framework} shows the proposed framework. All prompts used are shown in Appendix Table \ref{tab:all_prompts}.
To start the game, system prompts are used to assign different roles to LLM agents. Each system prompt for a role $p_i$ includes several important components: Role Information $\mathcal{RI}^{p_i}$ (Role Name and Role Introduction), Goal $\mathcal{G}^{p_i}$ (Winning Conditions), and Abstracted Strategy $\mathcal{S}^{p_i}$ for gameplay. The Role Name and Role Introduction provide information about the assigned role to the LLM agent, while the Goal (Winning Conditions) offers insights into how to achieve victory. Additionally, the Initial Playing Strategy outlines the high-level planning for the LLM agent to take specific actions during gameplay.

Below is a specific example of a system prompt for the role of Margana:

\noindent\textbf{Role:} \textit{Morgana}.

\noindent\textbf{Role Introduction:} \textit{In identification phase, you can identify teammates and the Assassin}.

\noindent\textbf{Goal:} \textit{Win the game by intentionally causing quests to fail for three rounds, alone or with teammates}.

\noindent\textbf{Initial Strategy:} 
\textit{You always pretend to be a loyal servant and recommend yourself as a candidate for quests, and let the quests fail}.

\subsection{Memory Storage}
Analyzing game history is vital for agents to grasp the current situation and make decisions. Yet, in Avalon, LLM agents' history responses are often too lengthy, surpassing input limits and potentially lowering performance. To tackle this, a memory storage system is introduced to record conversations among LLM agents, enabling subsequent analysis and decision-making.

\noindent\textbf{Memory Storage.}
Memory storage is vital for recording agents' conversation history in the current game round. It comprises structured memory objects containing key details like role name, detailed natural language responses, round number, and a flag indicating public or private status. Public information is visible to all roles, while private information pertains to each role's conversation. We assign separate memory pools to each agent for clarity in information processing. By storing this data, memory storage enables agents to access and review past conversations, improving their understanding of the game's progress.

% \textbf{Memory Updating: Summarization.}
\subsection{Memory Summarization.}
To store more information in memory, we use a summarization prompt to compress the information from the previous round and capture the essential details. The process of updating the memory with a summary of the previous round is illustrated below:

\begin{equation}
\mathcal{M}_t=\left\langle \operatorname{SMR}(\mathcal{M}_{t-1}),\left(\mathcal{R}^{p_1}_t \cdots, \mathcal{R}^{p_6}_t, \mathcal{I}_t\right)\right\rangle.
\end{equation}
The memory on round $t$ is $\mathcal{M}_t$. The response generated by the LLM for role $p_i$ on round $t$ is $\mathcal{R}^{p_i}_t$, and $\mathcal{I}_t$ represents the instructions and statements of the host on round $t$. $\left\langle \right\rangle$ is Text concatenation. $\operatorname{SMR}(\cdot)$ is the summarization prompting. 
% shown in Appendix Table \ref{tab:all_prompts}.

% The memory on round $t$ is represented as $\mathcal{M}_t$. The memory object containing the response generated by the LLM for role $p_i$ on round $t$ is denoted as $\mathcal{R}^{p_i}_t$, and $\mathcal{I}_t$ represents the instructions and statements of the host on round $t$. $\left\langle \right\rangle$ denotes text concatenation.
% $\operatorname{Summarize}(\cdot)$ is the summarization function based on the summarization prompting approach. The specific summarization prompt is shown in Table \ref{tab:all_prompts}.

\subsection{Analysis}
To help LLM agents improve strategic planning and increase their chances of winning, we introduce an analysis module. This module analyzes the role identity and potential strategies of other players during gameplay:
\begin{equation}
\mathcal{H}^{p_i}_t=\operatorname{ANA}\left(\mathcal{M}_t, \mathcal{RI}^{p_i}\right),
\end{equation}
where $\mathcal{M}_t$ is the memory on round $t$ and $\mathcal{RI}^{p_i}$ is the role information. By analyzing, LLM agents can better understand their collaborators and competitors, leading to improved decision-making and effective counterstrategies for winning.
% where $\mathcal{M}_t$ represents the memory on round $t$, and role information $\mathcal{RI}^{p_i}$ is provided by the system prompt.
% By providing this comprehensive analysis, LLM agents will be equipped with a deeper understanding of their collaborator and competitors, enabling them to make better decisions and develop effective counterstrategies that help lead to winning. 
% The example prompt is illustrated in Table \ref{tab:all_prompts}.

\subsection{Planning}
Agents need to understand the game progress and necessary strategies to win. Thus, a planning module is designed to create a strategic plan. The plan is based on the memory and information from the current round of the game, as described below:
\begin{equation}
\mathcal{P}^{p_i}_t=\operatorname{PLAN}\left(\mathcal{M}_t, \mathcal{H}^{p_i}_t, \mathcal{P}^{p_i}_{t-1}, 
    \mathcal{RI}^{p_i},
    \mathcal{G}^{p_i},
    \mathcal{S}^{p_i}\right),
\end{equation}
where $\mathcal{P}^{p_i}_t$ represents the strategic plan of agent ${p_i}$ at round $t$. $\mathcal{G}^{p_i}$ and $\mathcal{S}^{p_i}$ are goals and initial strategies.
By creating a strategic plan, the agents can have a flexible strategy for different situations. This foresight helps them make better decisions about collaborating with teammates, deceiving opponents, taking on the opposing faction's identity, and, if needed, sacrificing teammates or oneself to secure winning in the game.

\subsection{Action}
% \subsubsection{Action.}
In the action module, agents decide their next action based on memory information, situation analysis, and the strategic plan. There are five types of actions: selecting players, voting (agree or disagree), completing quests (succeed or fail), using non-verbal signals (raising hands, putting hands down, opening or closing eyes), and choosing to remain silent. The process of choosing the next action is as follows:
% Within the action module, agents will determine their next course of action by evaluating available memory information, conducting a thorough analysis, and adhering to the established strategic plan. There are five types of actions the agents can take including choosing players, voting (agree or disagree), performing quests (make quest succeed or fail), using non-verbal signals (raising hands up, put hands down, open eyes or close eyes), and choosing to remain silent. The process of choosing the next action with available information is illustrated below:
\begin{equation}
    \mathcal{A}^{p_i}_t \sim p\left( \mathcal{A} | \mathcal{M}_t, \mathcal{H}^{p_i}_t, \mathcal{P}^{p_i}_{t}, \mathcal{RI}^{p_i}, \mathcal{G}^{p_i},
    \mathcal{S}^{p_i}, \mathcal{I}'_t\right).
\end{equation}
The subsequent action depends on the memory, the comprehensive analysis, the strategic plan, and the instruction from the host. The details of these action decisions are confidential and only known to the respective agent. The host and other players cannot see these decisions.
% The determination of the subsequent action is contingent upon the utilization of the memory $\mathcal{M}_t$, the comprehensive analysis $\mathcal{H}^{p_i}_t$, the strategic plan $\mathcal{P}^{p_i}_{t}$, and the specific instruction provided by the host in round $t$. Within the action module, agents select a particular action in alignment with the established strategic plan. It is imperative to note that the details of these action decisions remain confidential and are accessible exclusively to the respective agent. Neither the host nor other players have visibility into these decisions.

\subsection{Response Generation}
The Response Generation module is responsible for generating a response to the host's inquiry. Agents in this module choose an action and provide an explanation to the host. Agents are given the freedom to collaborate, deceive, and assume the identity of the opposite faction in their explanations.
% Following meticulous analysis and decision-making procedures, the response to the host's inquiry is generated within the Response Generation module. Within this response, agents delineate the selected action and provide a corresponding explanation to the host. It is important to note that in crafting these explanations, agents are not bound by truthfulness. Consequently, agents are afforded the flexibility to collaborate with teammates, deceive opponents, and assume the identity of the opposite faction through natural language descriptions.

\subsection{Experience Learning}
In practical scenarios, players can improve their Avalon gameplay strategy through experience. They gain insights not only from their own perspective but also by observing other players' strategies. An ideal Avalon LLM agent should learn from both its own experiences and those of other players.

\subsubsection{Self-Role Strategy Learning}
In Step 1, agents generate three strategic recommendations for a player's role-specific gameplay in Avalon games based on the game history. Agents avoid mentioning specific players and instead use role names to make the suggestions applicable in future games.
In Step 2, agents enhance their strategies by incorporating the gathered suggestions while maintaining the original strategy's strengths.

\subsubsection{Other-Role Strategy Learning}
Avalon LLM agents summarize the strategies adopted by other players to facilitate learning from the strategies employed by other players. Prompts for the above steps are shown in Appendix Table \ref{tab:strategy_prompts}.

\section{Experiment}

\subsection{Implementation Details}

We developed the Avalon game program in Python, using the gpt-3.5-turbo-16k model as both our backend and the baseline's. In all experiments, we set the agent model's temperature to 0.3 and the LLM extractor's to 0. The number of suggestions generated for updating strategies is 3.
Game rules and role descriptions were set according to the baseline template \cite{xu2023exploring}, which leverages historical context, enhances agent reasoning, and learns from past mistakes. Detailed descriptions are provided in Section \ref{sec:appendix_gamerule}.

% For broader validation, we implemented our framework on the Llama2-7b-chat-hf model. However, LLaMA-based agents face constraints due to the model's language understanding capabilities and token limitations. Preliminary exploration without further analysis is discussed in Section \ref{sec:appendix_llama}.

\subsection{Evaluation Metrics}
We evaluate the performance of our framework based on metrics from two perspectives.

\subsubsection{Gameplay Outcome and Strategy.} From this perspective, we use metrics associated with the gameplay outcome and strategies to quantitatively evaluate the performance of the proposed agents and the baseline agents.

\noindent\textbf{Winning Rate (WR).} 
The winning rate is the percentage of games won out of the total played, calculated by dividing the number of wins by the total games played:
% The winning rate typically refers to the percentage of games a particular player or team wins out of the total number of games played. It's simply the number of wins divided by the total number of games played, expressed as following:
\begin{equation}
    WR = (\frac{\# Wins}{ \# Games\ Played})\times 100\%
\end{equation}

\noindent\textbf{Quest Engagement Rate (QER).} "Quest engagement rate" is the ratio of rounds a player joins the quest team to the total rounds played in the games. It's calculated as follows:
% The term "quest engagement rate" denotes the proportion of rounds in which a specific player participates in the quest team out of the total number of rounds played during the games. The calculation of the quest engagement rate (QER) is depicted by the following equation: 
% \leicomment{Higher is better or lower is better.}
\begin{equation}
    QER = (\frac{\# Engagement\ Rounds}{ \# Rounds})\times 100\%
\end{equation}

\noindent\textbf{Failure Vote Rate (FVR)} The quest result relies on success or failure cards from team members. The failure vote rate indicates the percentage of votes against quest success, calculated as follows:
% The quest outcome is determined by the success or failure cards submitted by the quest team members. The failure vote rate refer the percentage of votes against the success of a quest that leads to its failure. The failure vote rate is computed with the following equation:
\begin{equation}
    FVR = (\frac{\# Failure\ Votes}{ \# Votes}) \times 100\%
\end{equation}

% To evaluate the correctness of responses of agents to the questions of the moderator in the game, we adopt the Valid Response Rate.
% \noindent\textbf{Valid Response Rate (VRR)}. Agents are required to engage in discussion, select players, and vote. A Valid Response is defined as a response that adheres to these requirements. the VRR is calculated as follows:
% % The quest outcome is determined by the success or failure cards submitted by the quest team members. The failure vote rate refer the percentage of votes against the success of a quest that leads to its failure. The failure vote rate is computed with the following equation:
% \begin{equation}
%     VRR = (\frac{\# Valid\ Responses}{ \# Total\ Responses}) \times 100\%
% \end{equation}

\subsubsection{Social Behaviors.} 
From this perspective, we use ChatGPT to assist the analysis on the social behaviors of agents.

\noindent\textbf{Leadership.} We gauge AI agents' leadership using "Leader Approval Rate (LAR)". LAR is calculated by dividing total approval votes by total leader votes across 20 Avalon games. It reflects consensus among players on proposed quest teams.

\noindent\textbf{Persuasion.} 
To evaluate LLM agents' persuasion, we track two metrics: self-recommendation rate (proposing oneself for quests) and success rate (self-recommendation for quest participation).
%To assess LLM agents' persuasion skills, we use two metrics: self-recommendation rate and success rate. The self-recommendation rate is calculated by dividing the number of rounds where agents propose themselves for the quest team by the total rounds played. The success rate represents the proportion of rounds where self-recommendation leads to active participation in the quest team. ChatGPT is used to detect instances of LLM agents' self-recommendation and determine its success. 

\noindent\textbf{Camouflage.}
Detecting camouflage in AI agents is challenging. We focus on identifying instances where agents assume different identities in the initial round of each game. Behaviors include Self-Disclosure, Camouflage, and Withholding Identity.

\noindent\textbf{Teamwork and Confrontation.}We use ChatGPT to analyze role responses, aiming to identify instances of collaboration or confrontation. ChatGPT prompts with a player's response and evaluates trust (teamwork), lack of trust (confrontation), or ambivalence towards others.
%We conduct an evaluation focusing on the teamwork and confrontation rates among different roles. 
% We utilize ChatGPT to scrutinize the responses of various roles, with the specific objective of identifying instances where agents either collaborate with their fellow players (teamwork) or engage in confrontations with others (confrontation). This analysis is achieved by prompting ChatGPT with a specific player's response and evaluating the trust (teamwork), lack of trust (confrontation), or ambivalence towards other players. 

\noindent\textbf{Sharing.}
% In evaluating AI agents' collaborative dynamics, we introduce the metric "sharing" to measure their inclination to distribute vital information among teammates. Sharing reflects how often agents disclose valuable information, crucial for team cooperation. Using ChatGPT, we analyze agents' dialogues to identify instances of sharing behavior, aiming to quantify their willingness to share for the team's benefit.
Sharing reflects how often agents disclose valuable information, crucial for team cooperation. Using ChatGPT, we analyze agents' dialogues to identify instances of sharing behavior, aiming to quantify their willingness to share for the team's benefit.

\begin{table}
    \centering
    \resizebox{0.44\textwidth}{!}{ 
    \begin{tabular}{l|ccc}
    \toprule
        \textbf{Method}& \textbf{Good Side} & \textbf{Evil Side}\\ 
       \midrule
       Ours  & 90 & 100 \\
       \midrule
       \quad w/o analysis  & 60 & 60\\ 
       \quad w/o plan  & 80&100\\
       \quad w/o action  & 100 & 80\\
       \quad w/o strategy learning  & 50 & 60\\\bottomrule
    \end{tabular}
    }
    \vspace{-5px}
    \caption{Results of the gameplay between ours and baseline. We present the winning rates (WR) of our method being good and evil sides.}
    \vspace{-10pt}
    \label{tab:main_result}
\end{table}

\begin{figure}
    \centering
    \includegraphics[width=\linewidth]{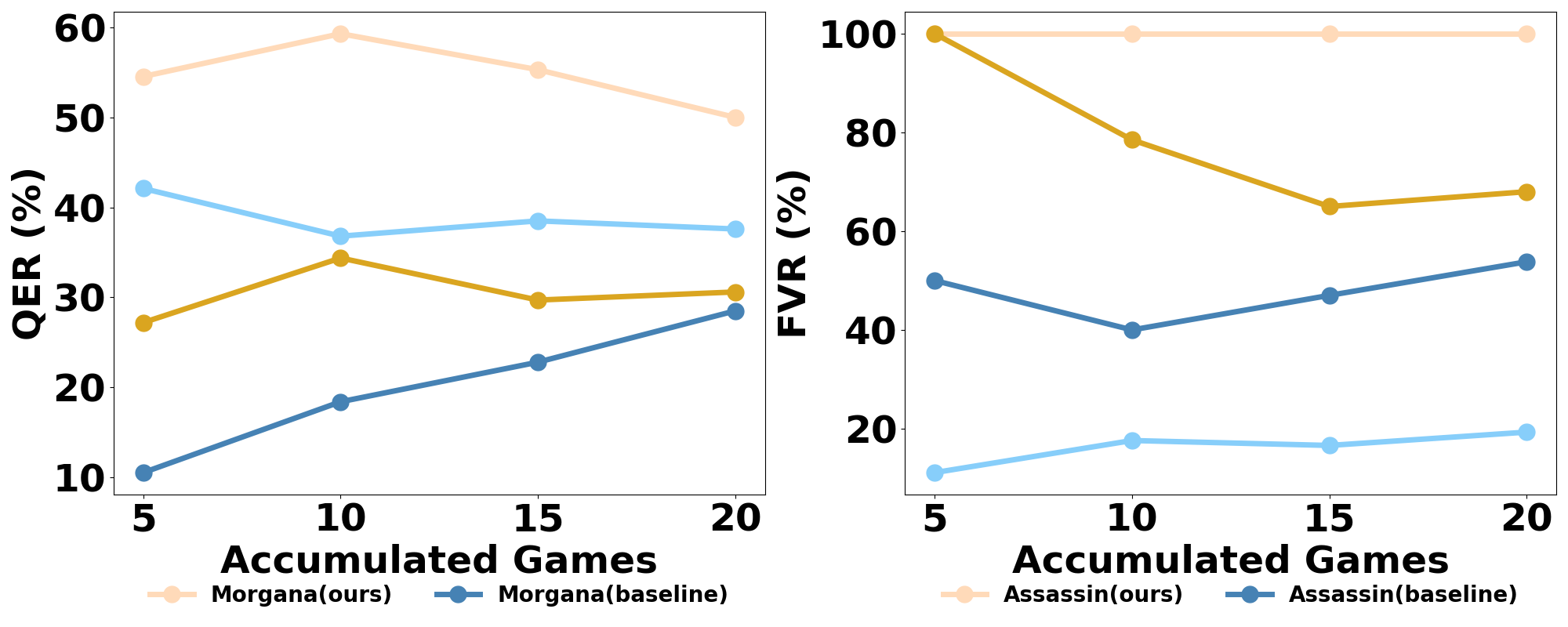}
    \vspace{-20pt}
    \caption{\textbf{(a)}: Comparison of the engaging quests rate when playing evil side. Higher engaging quests rate means more opportunities for the player to influence the outcome of the game. \textbf{(b)}: Comparison of the failure vote rate when playing evil side. Baseline is worse.}
    \vspace{-10pt}
    \label{fig:qer_fvr}
\end{figure}

\subsection{Experiment Results}
To validate the efficacy of Avalon AI agents, we repurposed Werewolf AI agents \cite{xu2023exploring} as baselines. Across two sets of 10 consecutive Avalon games, our agents faced off against the baselines, with Evil versus Good and vice versa. After the matches, we compared the winning rates of our Avalon AI agents to the baselines.
% To validate the efficacy of our proposed Avalon AI agents, we utilized Werewolf AI agents \cite{xu2023exploring} and adapted them for Avalon games as baseline agents. We conducted two series of 20 consecutive Avalon games, where our Evil Side played against the baseline Good Side for 20 games, and conversely, our Good Side played against the baseline Evil Side for another 20 games. 
% Following the gameplay sessions, we conducted a comparative analysis of the winning rates achieved by the proposed Avalon AI agents in contrast to the baseline. 
% \leicomment{What is the baseline? Missing the details of baseline.}
% As depicted in Figure \ref{fig:main_result}, our method demonstrated a 100\% winning rate in 20 games when playing the good side. Conversely, when playing the evil side, the winning rate was 70\% over the same number of games.
% Furthermore, an observation was made regarding a decrease in the winning rate, transitioning from 100\% to 66.6\%, as the number of consecutive Avalon games increased from 5 to 15. This decline may be attributed to the learning mechanism employed by the baseline, where responses from winning games carry a high probability of being utilized to enhance future gameplay. Consequently, the proposed Avalon AI agents engaged in competitive matches against the baseline agents. Notably, our Avalon Agents demonstrated an increase in the winning rate from 66.6\% to 70\% between games 15 and 20. 
As depicted in Table \ref{tab:main_result}, our method demonstrated a 90\% winning rate in 10 games when playing the good side. Conversely, when playing the evil side, the winning rate was 100\% over the same number of games.

% Furthermore, ablation studies are conducted to demonstrate the impact of key modules of our AI agents. Without the analysis module, our method's winning rates in both good and evil sides dropped to 60\%. This indicates the significant impact of the analysis module on game results, potentially helping our agents better understand the game situation and the identities of other players, thus making wiser decisions. 

% Ablation studies demonstrate the significance of key modules in our AI agents. Removing the analysis module decreased our method's winning rates to 60\% for both sides, highlighting its impact on our agents' understanding and decision-making. Excluding the planning module reduced the good side's winning rate to 80\%, emphasizing its role in devising effective strategies. Without the action module, the good side's winning rate reached 100\%, while the evil side's dropped to 80\%, underscoring its importance for the evil side's success. Removal of the strategy learning module resulted in both sides' winning rates decreasing to 50\% and 60\% for good and evil respectively, emphasizing its role in enhancing strategies and improving chances of victory. In conclusion, the analysis and strategy learning modules play a crucial role in determining game outcomes, with notable effects on the winning rates of both sides. Additionally, the planning and action modules are essential for success, given their significant impact on various aspects of gameplay.
Ablation studies reveal the importance of key modules in our AI agents. Removing the analysis module lowered winning rates to 60\% for both sides, showing its impact on understanding and decision-making. Excluding the planning module reduced the good side's winning rate to 80\%, highlighting its role in devising strategies. Without the action module, the good side won 100\% while the evil side dropped to 80\%, indicating its importance for the evil side's success. Removal of the strategy learning module led to winning rates decreasing to 50\% and 60\% for good and evil respectively, emphasizing its role in enhancing strategies. In conclusion, the analysis and strategy learning modules significantly influence game outcomes, affecting both sides' winning rates. Additionally, the planning and action modules are crucial for success, given their impact on gameplay.
To better grasp the strategies employed by our Avalon Agents and the baseline agent, we compared quest engagement and failure voting rates when different AI agents acted as the evil side. Both rates significantly impact game outcomes. A higher quest engagement rate allows more chances for players to influence the game, while a higher failure voting rate suggests a greater chance for the evil side to win but also increases the risk of exposure, indicating an aggressive gameplay approach.
% The outcomes regarding the quest engagement rate and the failure voting rate are presented in Figure \ref{fig:qer_fvr}. It is evident that our proposed AI agents adopt an assertive approach when assuming the roles of Morgana and Assassin, with an average quest engagement rate of 35.2\% and a failure voting rate of 62.3\% when joining the quest team. In contrast, the baseline agents exhibit a quest engagement rate of only 20.2\% and a relatively low failure voting rate of 26.2\%. Consequently, our proposed Avalon AI agents achieve a 70\% winning rate against the baseline agents when playing as the evil side.
% The outcomes regarding the quest engagement rate and the failure voting rate are presented in Figure \ref{fig:qer_fvr}. It is evident that our proposed AI agents adopt an assertive approach when assuming the roles of Morgana and Assassin, with an average quest engagement rate of 40.3\% and a failure voting rate of 84.0\% when joining the quest team. In contrast, the baseline agents exhibit a quest engagement rate of only 33.1\% and a relatively low failure voting rate of 36.5\%. Consequently, our proposed Avalon AI agents achieve a 90\% winning rate against the baseline agents when playing as the evil side.
Figure \ref{fig:qer_fvr} illustrates the outcomes for quest engagement and failure voting rates. Our AI agents, particularly when playing as Morgana and Assassin, show assertiveness, with a 40.3\% quest engagement rate and 84.0\% failure voting rate. In comparison, baseline agents have lower rates at 33.1\% and 36.5\% respectively. As a result, our proposed Avalon AI agents achieve a 100\% win rate against the baseline agents when playing as the evil side.

\begin{figure}[!htp]
    \centering
    % \vspace{-5pt}
    \includegraphics[width=\linewidth]{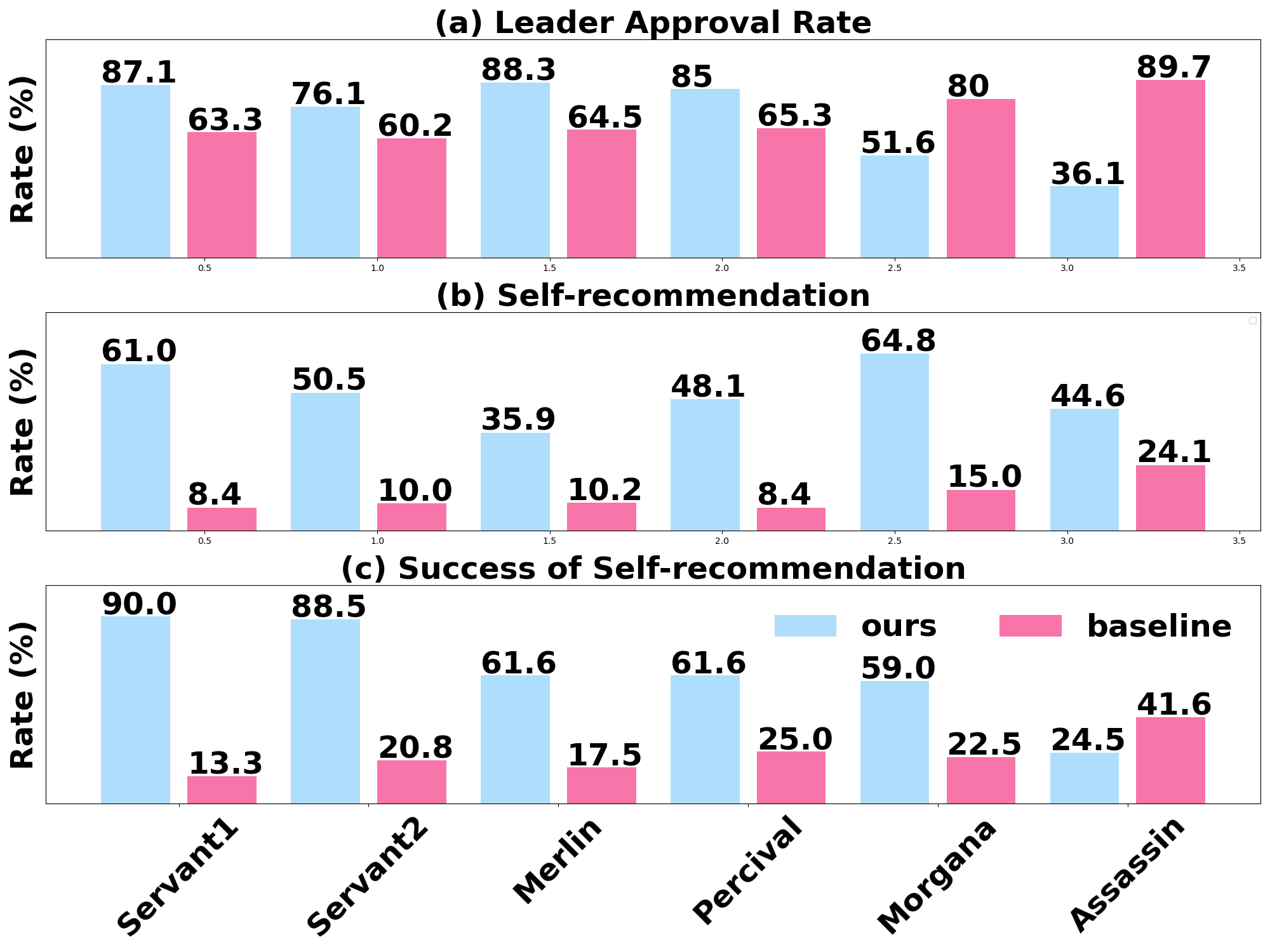}
    \vspace{-20px}
    \caption{(a): The leadership behavior. Players with higher Leader Approval Rate get more agreements from other players when deciding a quest team. (b) and (c): The persuasion behavior. Self-recommendation Rate: players with higher Self-recommendation Rate are more will to engage in quests. Self-recommendation Success Rate: players more likely to gain the trust of other players has higher Self-recommendation Success Rate. }
    \vspace{-5pt}
    \label{fig:leadership_and_persuasion}
\end{figure}

\section{Social Behaviors of AI Agents}

To evaluate if AI agents replicate human social behaviors in Avalon, we conduct a thorough analysis. This involves assessing the agents' execution of teamwork, leadership, persuasion, camouflage, and confrontation through the frequency distribution in game logs from two sets of 10 consecutive games.

% \begin{figure}
%     \centering
%     \includegraphics[width=\linewidth]{figures/persuasion.png}
%     \vspace{-10pt}
%     \caption{The persuasion behavior when playing different roles: Self-recommendation Rate: players with higher Self-recommendation Rate are more will to engage in quests. Self-recommendation Success Rate: players more likely to gain the trust of other players has higher Self-recommendation Success Rate.}
%     \vspace{-5pt}
%     \label{fig:persuasion}
% \end{figure}

\begin{figure}
    \centering
    \includegraphics[width=\linewidth]{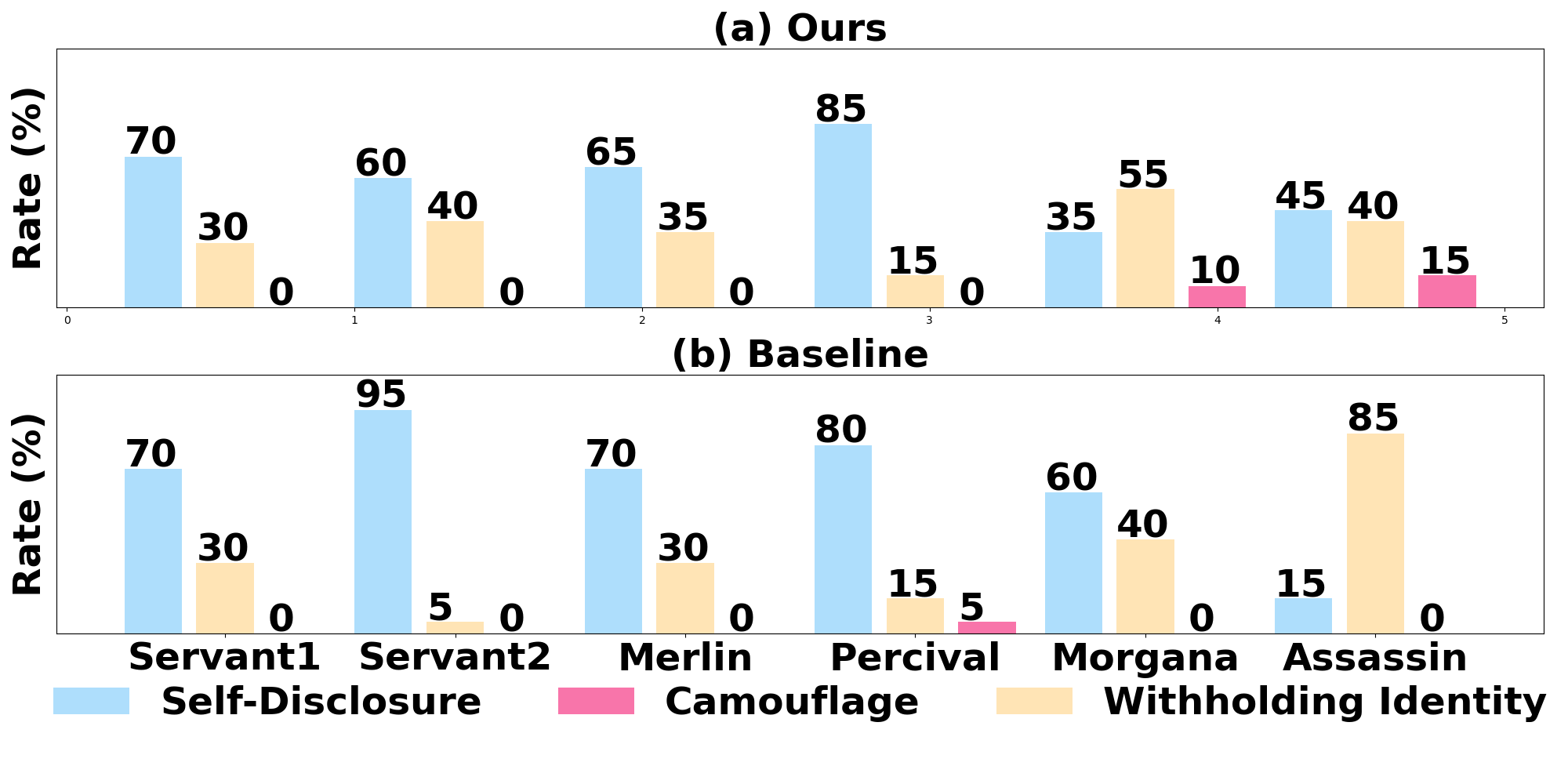}
    \vspace{-15px}
    \caption{The camouflage behavior when playing different roles: at first round of each game, the distribution of the players choose Self-Disclosure, Camouflage or Withholding Identity.}
    \label{fig:deception}
    \vspace{-10pt}
\end{figure}

% \begin{figure*}
%     \centering
%     \includegraphics[width=0.95\linewidth]{figures/teamwork_confrontation_ours.png}
%     \vspace{-15pt}
%     \caption{The teamwork and confrontation behaviors when playing different roles (Ours): each subfigure shows the attitude distribution of the player portraying specific role (on the top) towards players in other roles (on the left). }
%     \vspace{-15pt}
%     \label{fig:teamwork_confrontation_ours}
% \end{figure*}

% \begin{figure*}
%     \centering
%     \includegraphics[width=0.95\linewidth]{figures/teamwork_confrontation_baseline.png}
%     \vspace{-15pt}
%     \caption{The teamwork and confrontation behaviors when playing different roles (Baseline): each subfigure shows the attitude distribution of the player portraying specific role (on the top) towards players in other roles (on the left).}
%     \vspace{-15pt}
%     \label{fig:teamwork_confrontation_baseline}
% \end{figure*}

% \begin{figure*}
%     \centering
%     \includegraphics[width=0.95\linewidth]{figures/teamwork_confrontation_ablation_wo_strategy.png}
%     \vspace{-15pt}
%     \caption{The teamwork and confrontation behaviors when playing different roles (agents without experience learning module)}
%     \vspace{-15pt}
%     \label{fig:teamwork_confrontation_ablation_wo_strategy}
% \end{figure*}
\begin{figure*}
    \centering
    \includegraphics[width=\linewidth]{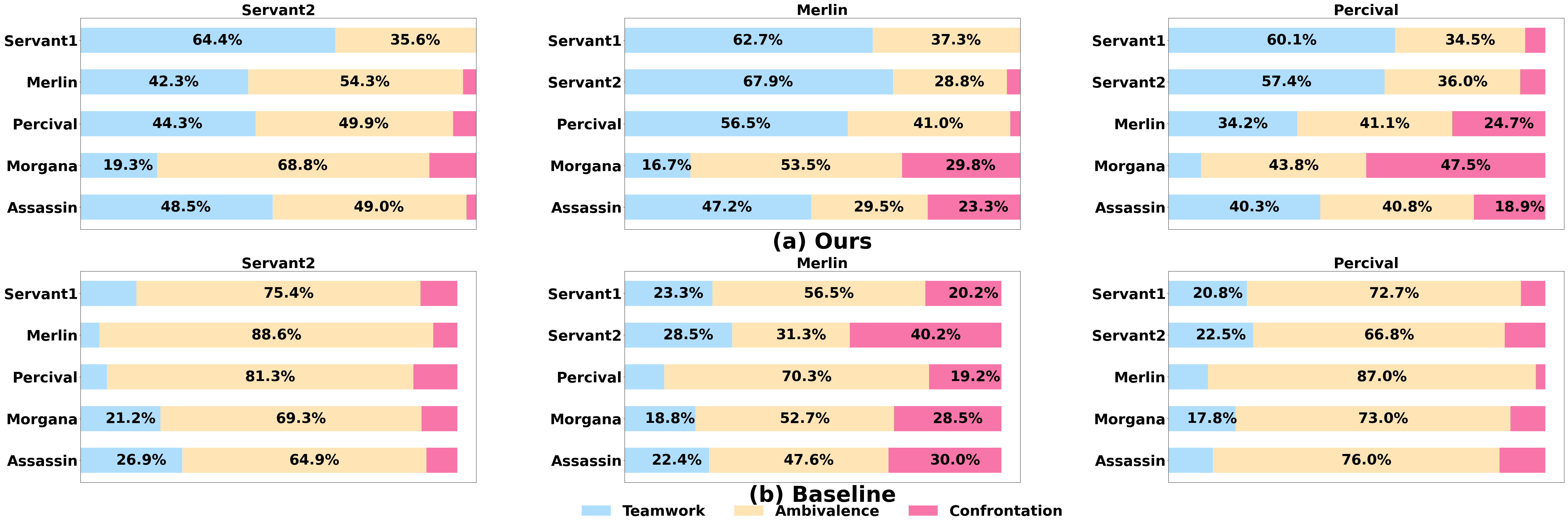}
    \vspace{-15pt}
    \caption{The teamwork and confrontation behaviors when playing different roles. Each subfigure shows the attitude distribution of the player portraying specific role (on the top) towards players in other roles (on the left).}
    \vspace{-10pt}
    \label{fig:teamwork_confrontation}
\end{figure*}

\begin{figure}
    \centering
    \includegraphics[width=\linewidth]{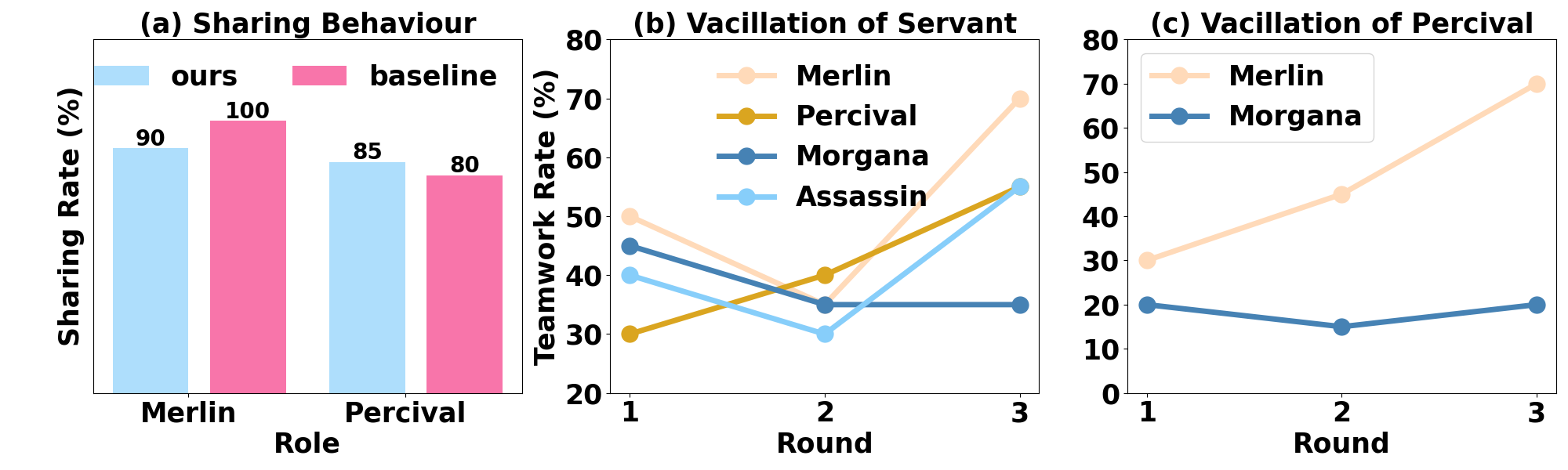}
    \vspace{-15pt}
    \caption{(a): The sharing behavior when playing Percival and Merlin at the first round. (b) and (c): The teamwork vacillation between different rounds.}
    \vspace{-15pt}
    \label{fig:share_and_vacillation}
\end{figure}

% \begin{figure*}
%     \centering
%     \includegraphics[width=\linewidth]{figures/teamwork_confrontation_demo.png}
%     \vspace{-10pt}
%     \caption{The teamwork and confrontation behaviors when playing different roles (agents without experience learning module)}
%     \vspace{-10pt}
%     \label{fig:tc}
% \end{figure*}

% \usepackage{subcaption}

% \begin{figure*}
%     \centering
%     \begin{subfigure}{0.4\linewidth}
%         \centering
%         \includegraphics[width=\linewidth]{figures/share.png}
%         \caption{子图 A}
%         \label{fig:subfigA}
%     \end{subfigure}
%     \begin{subfigure}{0.4\linewidth}
%         \centering
%         \includegraphics[width=\linewidth]{figures/vacillation.png}
%         \caption{子图 B}
%         \label{fig:subfigB}
%     \end{subfigure}
%     \caption{示例图表包含两个子图}
%     \label{fig:example}
% \end{figure*}
% \begin{figure*}
%     \centering
%     \subcaptionbox{子图 A\label{fig:subfigA}}{\includegraphics[width=0.4\linewidth]{figures/share.png}}
%     \hspace{0.01\linewidth} % 控制子图间的水平间距
%     \subcaptionbox{子图 B\label{fig:subfigB}}{\includegraphics[width=0.4\linewidth]{figures/vacillation.png}}
%     \caption{示例图表包含两个子图}
%     \label{fig:example}
% \end{figure*}

\subsection{Leadership} 

Leadership skills come into play when players take charge of discussions and decision-making processes. A good leader can steer the conversation, guide suspicions, and rally the loyal servants to make informed decisions. Leadership abilities are crucial for the good side to effectively counter the deceptive tactics employed by the evil side.

Figure \ref{fig:leadership_and_persuasion} (a) illustrates the Leader Approval Rate when agents assume various roles. It is evident that our agents, playing on the good side, attain remarkably high Leader Approval Rates when serving as leaders. Notably, the AI agents achieve a Leader Approval Rate exceeding 80\% averagely while undertaking roles associated with the good side. This signifies their robust leadership qualities and their proactive approach to steering the gameplay towards victory. However, the baseline agents could propose good side players to the quest team to achieve high Leader Approval Rate but low game win rate.

\subsection{Persuasion}
Figure \ref{fig:leadership_and_persuasion} displays the evaluation outcomes assessing the AI agents' persuasion ability. Notably, agents employ distinct strategies based on their assumed roles, as shown in Figure \ref{fig:leadership_and_persuasion} (b). When playing as Loyal Servant and Morgana, agents display a high self-recommendation rate for quest team participation, impacting mission success. Conversely, a cautious approach is seen with roles like Merlin, Percival, and Assassin, evident from their low self-recommendation rates. This strategic restraint is crucial, particularly for roles like Merlin, emphasizing the importance of concealing identity. From Figure \ref{fig:leadership_and_persuasion} (c), Loyal Servants exhibit higher success rates in self-recommendation compared to roles that easily raise suspicion. Additionally, the proposed Avalon Agents show higher rates of self-recommendation and greater success compared to baseline agents, indicating enhanced persuasion abilities.

\subsection{Camouflage}
Camouflage is central to Avalon. Evil roles must deceive loyal servants while subtly sabotaging missions. Skilled players create elaborate lies and misdirection. Loyal servants also engage in camouflage to conceal their identities, especially when under suspicion.

In Figure \ref{fig:deception}, the rates of various behaviors exhibited by AI agents are displayed. Notably, the agents display a notably high tendency to reveal their identities at the commencement of the game, particularly among the roles associated with the good side. Intriguingly, in the roles of Morgana and Assassin, agents opt to either conceal or assume different identities without explicit instructions to do so in the initial strategy. Specifically, Morgana and the Assassin display rates of assuming alternate identities of 10\% and 15\%, respectively, a strategy akin to that observed in human players, where Percival perceives both Merlin and Morgana but lacks precise knowledge of their identities. This spontaneous adoption of deceptive behaviors by AI agents stands out as a captivating observation, underscoring their adaptability and strategic acumen in the pursuit of game victory.

\subsection{Teamwork and Confrontation}
Teamwork is vital for loyal servants to identify each other and succeed in missions by strategizing, discussing assignments, and sharing information to uncover evil roles. Confrontations arise when suspicions lead to accusations, resulting in intense exchanges where accusers present reasoning and the accused offer defenses or deflect suspicion onto others.

In Figure \ref{fig:teamwork_confrontation} (a), teamwork and confrontation rates of good side roles are depicted. Loyal Servants tend to avoid confrontation due to their lack of specific identity information. However, Merlin, aware of Morgana and Assassin, confronts them frequently. Percival, aware of Merlin and Morgana without knowing their exact identities, confronts both. These observations highlight the adaptive strategies of AI agents, mirroring the social dynamics of human players in Avalon.

Figure \ref{fig:teamwork_confrontation} (b) shows teamwork and confrontation rates of baseline agents. Rates remain consistent across roles, suggesting they do not adjust strategies based on role assumptions.

\subsection{Sharing}
Sharing is essential for Percival and Merlin. They possess more information than other good roles, and sharing their insights aids in winning the game. However, excessive sharing of known information may also benefit the opposing side, as discussions are public to all players. Therefore, strategic sharing of information is necessary to win the game.

Figure \ref{fig:share_and_vacillation} (a) depicts the proportion of known information shared with other players by different agents playing the roles of Merlin and Percival in the first round of the game. It is observed that both the agents designed by us and the baseline agents exhibit an excessive level of sharing behaviors.

\subsection{Vacillation}
At the game's onset, some players possess identity clues, like Percival knowing Morgana and Merlin without distinction, while others, like Loyal Servants, lack such info. Both situations require players to deduce identities for their camp's benefit. Analyzing teamwork proportions across rounds reveals players' ability to discern allies and foes.

Figure \ref{fig:share_and_vacillation} (b) illustrates Loyal Servants' teamwork tendencies, while (c) shows Percival's tendencies towards Morgana and Merlin. Throughout the game, players increasingly collaborate with teammates and less with enemies. However, Loyal Servants face greater challenges inferring roles, leading to higher teamwork with potential foes.

\subsection{Behavior Spontaneity}
Teamwork and confrontation behaviors of players arise spontaneously due to game mechanics fostering interaction and competition. Teamwork aids in identifying evil roles, facilitating successful quests. However, teamwork often brings confrontation, as doubts about role identities persist. Even without strategic learning mechanisms, players exhibit these behaviors, showing their spontaneous nature. However, behavior distributions vary significantly between agents with and without strategic learning. The relevant analysis is provided at the Section \ref{sec:appendix_teamwork_confrontation}.

\section{Conclusion}
This paper explores the social behaviors of LLM-based agents in the Avalon game. We introduce a multi-agent framework facilitating efficient communication and interaction. This framework includes memory, analysis, planning, action, and response modules capable of learning from experience. Unlike prior studies, our research delves into the social dynamics of these agents in gameplay scenarios. Our evaluation showcases the success of our framework in achieving winning strategies and the adaptability of LLM agents in complex social interactions. Future work involves optimizing our approach, exploring its applicability in diverse game environments, and further understanding LLM agents' potential in dynamic social interactions.

\section{Limitations}
Although the LLM agent framework we proposed has performed well in the Avalon game, there are also limitations of high cost and slow interaction speed, due to multiple accesses to the model required for each interaction. Additionally, from the behaviors exhibited by the agent, there are also instances of unreasonable behavior distribution, such as excessive self-disclosure actions. In the future, we will explore and improve these aspects.

\section*{Acknowledgements}
This research is supported, in part, by SMP-IDATA Open Youth Fund.
This research is supported, in part, by the National Key R\&D Program of China (Grant No.2023YFF0725001), National Natural Science Foundation of China (Grant No.92370204), Guangzhou-HKUST(GZ) Joint Funding Program (Grant No.2023A03J0008), Education Bureau of Guangzhou Municipality.

\bibliography{anthology,custom}
\bibliographystyle{acl_natbib}

\appendix
\clearpage

\section{Appendix}\label{sec:appendix}

\subsection{Avalon Introduction}
\label{avalon_intro}
% Avalon, also known as ``The Resistance'', is a social deduction board game. 
Avalon is designed for 5 to 10 players. Specifically, we focus on the 6-player variant of the game.
% This game is adapted for 5-10 players. 
% In our work, we adopt the 6-player version of the game.

\noindent\textbf{Player roles.} 
Roles including Merlin, Percival, Morgana, Assassin, and two Loyal Servants, are divided into good and evil sides. Merlin, Percival, and loyal servants are on the good side, while Morgana and Assassin are on the evil side. Players are assigned roles secretly, with some having special abilities. Morgana and Assassin are initially aware of each other. Percival is able to see Merlin and Morgana but does not know their exact identities. Merlin is aware of the identities on the evil side.
% The roles include Merlin, Percival, Morgana, Assassin and two Loyal Servants. These roles are divided into the good and evil sides. Among them, Merlin, Percival and loyal servants belong to the good side, while Morgana and Assassin belong to the evil side. Each player will be assigned a role secretly, where some roles may have special abilities. Specifically, Morgana and Assassin know each other at the beginning. Percival sees both Merlin and Morgana but does not know their exact identities. Merlin knows the identity of the evil characters, but they do not know who Merlin is. 

\noindent\textbf{Quest team assignment.}
After receiving roles, players engage in 3-5 rounds of discussion and voting for a certain number of players to form a quest team. At the start of each round, a leader is assigned in rotation. The leader hosts a discussion, followed by a public vote on quest team members. If more than half of the votes agree, the team forms; otherwise, leadership rotates to the next player for further discussion and voting. Each round allows up to five discussion and voting cycles, with the leader directly assigning team members after the fifth round.
% After players receiving their respective roles, they will engage in 3-5 rounds of discussion and vote process to assemble a quest team, which consists of 2 or 3 players. At the beginning of a round, a leader will be assigned on a rotational basis. The leader hosts the discussion, then all players need to vote on the quest team members in public. If more than half of the votes are in agreement, the quest team will be successfully assembled. Otherwise, the leader moves to the next player and the players discuss and vote again. In each round, discussion and vote process can be conducted up to five times. The leader can directly assign quest members after the fifth discussion. 

\noindent\textbf{Quest phase.} 
The quest outcome is determined by the cards submitted by the quest team. Good players can only submit success cards, while evil players can choose to submit either success or failure cards. A quest is successful if all team members vote for success, and fails if one or more members vote for failure.
% The quest outcome is determined by the success or failure cards submitted by the quest team members. Players on the good side can only submit success cards, while players on the evil side have the option to submit either success or failure cards. A quest is deemed successful if all team members vote for success, whereas it fails if one or more members cast a failure vote.
% If all quest team members vote to succeed, then the quest will succeed. If any member votes to fail, then the quest fails. 

\noindent\textbf{End of the game.} 
The game ends when three quests succeed (good side wins) or three quests fail (evil side wins). Additionally, the evil players can win by correctly identifying Merlin at the end.
% The game ends when either three quests succeed, resulting in victory of the good side, or three quests fail, leading to victory of the evil side. However, the evil players can still secure a win at the end by accurately identifying which player is Merlin.

\subsection{Game Rules and Role Description}\label{sec:appendix_gamerule}

\begin{quote}
    {\small
    \small\texttt{You are playing a game called the Avalon with some other players. This game is based on text conversations. Here are the game rules: \\
    Roles: The moderator is also host, he organised this game and you need to answer his instructions correctly. Don’t talk with the moderator. There are five roles in the game, Merlin, Percival, Loyal Servant, Morgana, Assassin. Merlin, Percival and Loyal Servant belong to good side and Morgana and Assassin belong to evil side There are two alternate phases in this game, reveal phase and quest phase. When it’s reveal phase: You need follow the instructions of the moderator. You needn’t worry about other players and moderator knowing what you say and do. No need to worry about suspicions from others during the phase. If you are Merlin, you can know which two players are Morgana and Assassin but you can't know which one is Morgana or Assassin specifically. If you are Percival, you can know which two players are Merlin and Morgana but you can't know which one is Merlin or Morgana specifically. If you are Morgana, you can know which player is Assassin.If you are Assassin, you can know which player is Morgana. If you are Loyal Servant, you can't get any information in this phase. \\
    The quest phase includes 5 rounds. A round includes discussion, voting and execute quest. At every round, all players need to discuss about which players (candidates) will execute the quest at current round. And then all players need to vote if the candidates should execute the quest, if the agreement exceeds 1/2, the candidates will execute the quest, otherwise, discuss again and vote again. When executing quest, the candidates need to choose to make quest successful or failed. If all candidates choose to make quest successful, the quest will succeed. If anyone makes the quest failed, the quest will fail. \\
    At the end of a round, if the quest succeed, good side will get one point, otherwise, evil side will get one point. Which side get 3 points earlier, which side wins the game. If you are Assassin, at the end of a round, you can choose to identify which one is Merlin, if the identifying is successful, the red camp directly win the game. If not successful, the Assassin will expose his identification. \\
    Objectives: your goal is to help your side get 3 points and win the game. If you are Assassin, you also need to reason which player is Merlin as early as possible. \\
    Tips: To complete the objective: you should analyze and use your ability correctly. During quest phase, you need to reason carefully about the roles of other players and be careful not to reveal your own role casually unless you’re cheating other players. Only give the player’s name when making a decision/vote, and don’t generate other players’ conversation. Reasoning based on facts you have observed and you cannot perceive information (such as acoustic info) other than text. You are \{player\}, the \{role\}. You’re playing with 5 other players. Do not pretend you are other players or the moderator. Always end your response with ‘<EOS>’. 
}
    }
    \end{quote}

\subsection{Module Prompts}
Our designed prompts for different modules are presented in Tables \ref{tab:all_prompts} and \ref{tab:strategy_prompts}.
% Table \ref{tab:all_prompts}

\subsection{Heuristic Rules for LLM Gameplay}
\begin{table*}
\centering
\resizebox{0.98\textwidth}{!}
{
\begin{tabular}{p{0.95\textwidth}}
\toprule
\large\textbf{Selection Case:}  \\
\hangindent=1em \hangafter=1 \qquad \small\texttt{Please extract the answer to the question from the answer.}\\
\hangindent=1em \hangafter=1 \qquad \small\texttt{Question: Please start discussing the candidates for the turn 1 quest (please explain the reasons for your choice appropriately). The number of players in this quest is 2.}\\
\hangindent=1em \hangafter=1 \qquad \small\texttt{Answer: I suggest that for the turn 1 quest, we consider including both player 1 and player 2. Player 1 has consistently shown loyalty and dedication, and player 2 strongly implied their role as Morgana during the reveal phase. By including both players, we can increase our chances of successfully completing the quest. Thank you.}\\
\large\textbf{Extractor:}  \\
\hangindent=1em \hangafter=1 \qquad \small\texttt{1, 2}\\
\midrule

\large\textbf{Voting Case:}  \\
\hangindent=1em \hangafter=1 \qquad  \small\texttt{Complete the following instructions based on the questions and answers. }\\
\hangindent=1em \hangafter=1 \qquad \small\texttt{Response true, if the answer expresses: 1.obvious agreement to something. 2.obvious decision to do something. 3.obvious expression of making quest successful. 4.obvious yes to something}\\
\hangindent=1em \hangafter=1 \qquad \small\texttt{Response false, if the answer expresses: 1.obvious disagreement to something. 2.obvious decision not to do something. 3.obvious expression of making quest failed. 4.obvious no to something}\\
\hangindent=1em \hangafter=1 \qquad \small\texttt{Response none, if the answer is not related to the question or didn't express a obvious attitude to something. }\\
\hangindent=1em \hangafter=1 \qquad \small\texttt{Question: The quest leader decides that the player 1player 3, player 5 will engage in the quest. You can choose to agree or object. Please vote to agree or disagree.}\\
\hangindent=1em \hangafter=1 \qquad \small\texttt{Answer: I vote to disagree.}\\
\large\textbf{Extractor:}  \\
\hangindent=1em \hangafter=1 \qquad \small\texttt{False}\\
\bottomrule
\end{tabular}
}
\caption{Cases of LLM-based extraction}
\label{tab:extraction_case}
\end{table*}

In the gameplay, we used LLM to extract information from the responses of the agents. For example, when the agent selects a player, it extracts the player number, and when voting, it extracts the player's voting result. With several demonstrations of how to extract corresponding information, LLM can extract information very accurately to help the game proceed smoothly. Table \ref{tab:extraction_case} shows some cases of extraction. 

It is observed agents sometimes may fail to answer questions correctly, such as voting with unclear attitudes. In order to allow the game to proceed smoothly, we design the following heuristic rules. When voting for quest candidates, if the agent's answer is unclear, we assume that it agrees. When voting the quest for success or failure, if the agent's answer is unclear, we default to it voting for failure. When agents select an excessive number of players, we truncate the selection to meet the quest's requirements. In cases where the agents choose too few players, the host will repeat question to the agent. If the required player count is still not met even after multiple retries, the program steps in to assist by making a random selection on behalf of the agent.

\begin{table*}
\centering
\resizebox{0.98\textwidth}{!}
{
\begin{tabular}{p{0.95\textwidth}}
\toprule
\large\textbf{Summarization:}  \\
\hangindent=1em \hangafter=1 \qquad \small\texttt{Within the context of the Avalon game, please assist \{Player i\} in summarizing the conversations known to him from the current phase. These conversations are structured in JSON format, with ``message'' signifying the content of the conversation, "name" identifying the speaker, and ``message\_type'' indicating the type of message relevant to \{Player i\}. Specifically,``public'' implies that all players have access to the message, while ``private'' implies that only \{Player i\} has access to it.}\\
\hangindent=1em \hangafter=1 \qquad  \small\texttt{Conversations: \{conversations\}.} 
\\
\midrule

\large\textbf{Analysis:}  \\
\hangindent=1em \hangafter=1 \qquad  \small\texttt{Your task is to analyze roles and strategies of the players who might be your enemies according to their behaviors. 
The analysis should be no more than 100 words. 
The behaviors are summarized in paragraphs.}\\
\hangindent=1em \hangafter=1 \qquad \small\texttt{Your name is \{Name\} your role is \{Role\}.}\\
\hangindent=1em \hangafter=1 \qquad \small\texttt{The summary is \{Summary\}.}\\
\midrule

\large\textbf{Planning:}  \\
\hangindent=1em \hangafter=1 \qquad \small\texttt{Your task is to devise a playing plan that remains in harmony with your game goal and existing strategy, while also incorporating insights from your previous plan and current environment state.} \\
\hangindent=1em \hangafter=1 \qquad \small\texttt{\{Role Information\}} \\
\hangindent=1em \hangafter=1 \qquad \small\texttt{Goal: \{Goal\}} \\
\hangindent=1em \hangafter=1 \qquad \small\texttt{Strategy: \{Strategy\}} \\
\hangindent=1em \hangafter=1 \qquad \small\texttt{Your previous plan: \{Plan\}} \\
\hangindent=1em \hangafter=1 \qquad \small\texttt{Summary of previous rounds: \{Summary\}} \\
\hangindent=1em \hangafter=1 \qquad \small\texttt{Analysis about other players: \{Analysis\}}. \\
\midrule

\large\textbf{Action:}  \\
\hangindent=1em \hangafter=1 \qquad \small\texttt{Your objective is to make decisions based on your role, your game goal and the current game state. There are five types of actions you can take: choosing players, voting (agree or disagree), performing missions (make missions succeed or fail), using non-verbal signals (raise hands up, put hands down, open eyes, or close eyes), and choosing to remain silent. Only one action type can be selected at a time. If you decide to choose players, you can choose multiple players according to Host's question.} \\
    \hangindent=1em \hangafter=1 \qquad \small\texttt{\{Role Information\}}\\
    \hangindent=1em \hangafter=1 \qquad \small\texttt{Goal: \{Goal\}}\\
    \hangindent=1em \hangafter=1 \qquad \small\texttt{Strategy: \{Strategy\}}\\
    \hangindent=1em \hangafter=1 \qquad \small\texttt{Your current plan: \{Plan\}} \\
    \hangindent=1em \hangafter=1 \qquad \small\texttt{Summary of previous rounds: \{Summary\}}\\
    \hangindent=1em \hangafter=1 \qquad \small\texttt{Analysis about other players: \{Analysis\}}.\\
    \hangindent=1em \hangafter=1 \qquad \small\texttt{Host's Instruction: \{Instruction\}}.\\
\midrule

\large\textbf{Response:}  \\
\hangindent=1em \hangafter=1 \qquad\small\texttt{Your task is to provide detailed response to the question of Host, in accordance with the provided actions. Your response should be no more than 100 words.}\\
    \hangindent=1em \hangafter=1 \qquad \small\texttt{\{Role Information\}}\\
    \hangindent=1em \hangafter=1 \qquad \small\texttt{Goal: \{Goal\}}\\
    \hangindent=1em \hangafter=1 \qquad \small\texttt{Strategy: \{Strategy\}}\\
    \hangindent=1em \hangafter=1 \qquad \small\texttt{Your current plan: \{Plan\}} \\
    \hangindent=1em \hangafter=1 \qquad \small\texttt{Summary of previous rounds: \{Summary\}}\\
    \hangindent=1em \hangafter=1 \qquad \small\texttt{Host's Instruction: \{Instruction\}}.\\
     \hangindent=1em \hangafter=1 \qquad \small\texttt{current actions: \{actions\}}\\
\bottomrule
\end{tabular}
}
\caption{Input prompts of our proposed different modules.}
\label{tab:all_prompts}
\end{table*}

\begin{table*}
\centering
\resizebox{0.98\textwidth}{!}
{
\begin{tabular}{p{0.95\textwidth}}
\toprule
\large\textbf{Self-Role Strategy Learning (Step 1)} \\ 
\hangindent=1em \hangafter=1 \qquad \small\texttt{Your task is to provide 3 suggestions for \{player\}'s playing strategy of the role \{role\} in Avalon games, according to the game log. The game log includes the summaries of different rounds of a game.} \\
\hangindent=1em \hangafter=1 \qquad \small\texttt{The roles of the players: \{player-role mapping\}}\\
\hangindent=1em \hangafter=1 \qquad \small\texttt{The summaries of a round game: \{summary\}}\\
\hangindent=1em \hangafter=1 \qquad \small\texttt{\{player\}'s game goal: \{goal\}}\\
\hangindent=1em \hangafter=1 \qquad \small\texttt{\{player\}'s playing strategy of role \{role\}:\{current strategy\}}\\
\hangindent=1em \hangafter=1 \qquad \small\texttt{Previous suggestions: \{suggestions from last game\}} \\
\hangindent=1em \hangafter=1 \qquad \small\texttt{Give your suggestions, No more than two sentences per suggestion and the suggestions should be general for future games (This implies that you should avoid referencing player x directly and instead use the respective role names when making your suggestion.) and effectively help him achieve his game goal in future games.} \\

\large\textbf{Self-Role Strategy Learning (Step 2)} \\
\hangindent=1em \hangafter=1 \qquad \small\texttt{Your task is to help \{player\} improve his playing strategy of the role \{role\} a Avalon game with suggestions.}\\
    \hangindent=1em \hangafter=1 \qquad \small\texttt{\{player\}'s strategy: \{current strategy\}} \\
    \hangindent=1em \hangafter=1 \qquad \small\texttt{Suggestions: \{suggestions\}}\\
    \hangindent=1em \hangafter=1 \qquad \small\texttt{Please improve the strategy while retaining the advantages of the original strategy for him and the strategy should be no more than 2 sentences. Describe the strategy you provide using continuous sentences rather than bullet points or numbering.}  \\
\midrule
\large\textbf{Other-Role Strategy Learning}\\
\hangindent=1em \hangafter=1 \qquad \small\texttt{Your task is to help \{player\} analyze the strategies of other players in a Avalon game, according to the game log. The game log is summarized in paragraphs.} \\
    \hangindent=1em \hangafter=1 \qquad \small\texttt{The roles of the players: \{player-role mapping\}}\\
    \hangindent=1em \hangafter=1 \qquad \small\texttt{The summaries of rounds of the game: \{summary\}}\\
    \hangindent=1em \hangafter=1 \qquad \small\texttt{Previous strategies of other roles: \{previous strategies\}}\\
    \hangindent=1em \hangafter=1 \qquad \small\texttt{Your analysis should be no more than 100 words and the analysis should be general for future games (This implies that you should avoid referencing player x directly and instead use the respective role names when giving your analysis). And analyze together with previous strategies.}\\

    \hangindent=1em \hangafter=1 \qquad \small\texttt{For example:
    The strategy of Merlin is that ... The strategy of Assassin is that... The strategy of ... is ...}\\

\bottomrule
\end{tabular}
}
\caption{Input prompts of our experience learning module.}
\label{tab:strategy_prompts}
\end{table*}

\subsection{Ablation Study}

To validate the efficacy of the proposed modules, we conducted an ablation study under both with and without learning from experience setting. Initially, we assessed the effectiveness of the Improving Strategy Module (IS), the Analysis of Others' Strategies Module (AO), and the Analysis Module (AM) within the context of the learning from experience setting, wherein strategies were updated based on accumulated gameplay for both our agents and the baseline agents. In this evaluation, the proposed agents engaged in ten games, assuming evil side roles, against the baseline agents for each module. Following these games, the wining rate (WR), quest engagement rate (QER), and the failure voting rate (FVR) were measured and reported for analysis.
Table \ref{tab:ablation1} presents the outcomes of the ablation study conducted within the learning-from-experience setting. It is discernible that in the absence of the Improving Strategy module, where the strategy remains static but the agent can still glean insights from other players' strategies, the winning rate decreases by 20\%. Additionally, the agents exhibit reduced aggression, indicated by lower quest engagement rates and failure voting rates. Furthermore, the absence of the Analysis of Others' Strategies module and the Analysis Module also leads to a decline in the winning rate. In these scenarios, the agents adopt a cautious gameplay approach, resulting in significantly lower quest engagement rates but higher failure voting rates.
% To validate the efficacy of the proposed modules, we conducted an ablation study in both a learning-from-experience and a non-learning setting. Initially, we evaluated the Improving Strategy Module (IS), Analysis of Others' Strategies Module (AO), and Analysis Module (AM) within the learning-from-experience context. The agents engaged in ten games, assuming evil side roles, against baseline agents for each module. Win rate (WR), quest engagement rate (QER), and failure voting rate (FVR) were measured for analysis. Table \ref{tab:ablation1} presents the outcomes of the ablation study conducted within the learning-from-experience setting. Without the Improving Strategy module, the win rate decreased by 20\%, leading to reduced aggression and lower engagement in quests. The absence of the Analysis of Others' Strategies module and the Analysis Module also resulted in a decline in the win rate, with cautious gameplay and lower quest engagement rates but higher failure voting rates.

\begin{table}[h]
    \centering
    \resizebox{0.46\textwidth}{!}{ 
    \begin{tabular}{c|c|c|c|c|c}
        \toprule
        Method& WR(\%) & \multicolumn{2}{|c|}{QER(\%)}  & \multicolumn{2}{|c}{FVR(\%)} \\
        % \hline
        &&Morgana&Assassin&Morgana&Assassin\\
        \midrule
        full & 80 & 44.1& 49.1& 66.6& 78.5\\
        w/o. IS & 60 &42.8 &39.3& 46.1& 100  \\
        w/o. AO  & 70 &18.3 &8.3& 100& 100 \\
        w/o. AM & 50 &29.3 &39& 87.5& 100 \\
        \bottomrule
    \end{tabular}
    
    }
    \caption{Ablation Study on Experience Learning: Compare of full framework, without improving strategy (IS), without analysis strategies of others (AO) and without analysis module (AM).}
    \vspace{-15pt}
    \label{tab:ablation1}
\end{table}
% \begin{table}[h]
%     \centering
%     \resizebox{0.46\textwidth}{!}{ 
%     \begin{tabular}{c|c|c|c|c|c}
%         \toprule
%         Method& WR(\%)\\
%         % \hline
%         All Modules & \\
%         \midrule
%         \multicolumn{2}{}{Ablation on Experience Learning Module}\\
%         \midrule
%         w/o IS & 60 \\
%         w/o AO  & 70 \\
%         w/o AM & 50 \\
%         \midrule
%         \multicolumn{2}{}{Ablation on Analysis Module}\\
%         \midrule
%         teammates only & 90
%         adversaries only & 90
%         w/o analysis
%         \bottomrule
%     \end{tabular}
    
%     }
%     \caption{Ablation Study on Experience Learning: Compare of full framework, without improving strategy (IS), without analysis strategies of others (AO) and without analysis module (AM).}
%     \vspace{-15pt}
%     \label{tab:ablation1}
% \end{table}

Following the initial evaluation, we proceeded to assess the effectiveness of the Analysis Module, Planning Module, and Action Module under conditions where learning from experience was not incorporated. In this scenario, strategies were not updated for both our agents and the baseline agent. It is essential to note that the games were conducted independently, with no influence from previous games on future gameplay.
Table \ref{tab:ablation2} presents the results from the module ablation study conducted without incorporating learning from experience. It is discernible that the absence of the planning module results in a notable 20\% decrease in the winning rate. Additionally, the Assassin exhibits a significantly lower quest engagement rate, indicating a tendency to overlook the mission objective without the guidance of a strategic plan. This underscores the critical importance of the planning module in ensuring that agents consistently progress toward winning the game.Furthermore, in the absence of both the analysis and action modules, the agents exhibit a slightly lower quest engagement rate. Despite this, they manage to maintain an impressive 80\% winning rate. 

\begin{table}[]
    \centering
    \resizebox{0.46\textwidth}{!}{ 
    \begin{tabular}{c|c|c|c|c|c}
    \toprule
        Method& WR(\%) & \multicolumn{2}{|c|}{QER(\%)}  & \multicolumn{2}{|c}{FVR(\%)}\\
        &&Morgana&Assassin&Morgana&Assassin \\ \midrule
       all modules & 90 & 55.5& 58.3 &93.7 & 100 \\
       w/o analysis & 80 & 44.1 & 47.5 &100 &100\\ 
        w/o. plan  & 60&55 & 16.6& 90 &100\\
         w/o. action  & 80 &45.6 &45.6 & 100& 100\\\bottomrule
    \end{tabular}
    }
    \caption{Module Ablation: under the setting without learning from experience.}
    \label{tab:ablation2}
\end{table}

\begin{table}
    \centering
    % \vspace{-10pt}
    \resizebox{0.46\textwidth}{!}{ 
    \begin{tabular}{c|c|c|c|c|c}
    \toprule
        Method& WR(\%) & \multicolumn{2}{|c|}{QER(\%)}  & \multicolumn{2}{|c}{FVR(\%)}\\
        &&Morgana&Assassin&Morgana&Assassin\\
        \midrule
       all players & 90 & 55.5& 58.3 &93.7 & 100 \\
       teammates only & 80 & 26.8 & 48.1 &62.5 &100\\ 
       adversaries only  & 90& 38.3 & 45.3 & 92.3 &100\\
    \bottomrule
    \end{tabular}
        }
    \caption{Analysis Module Ablation: under the setting without learning from experience. Analyzing different objects.}
    % \vspace{-20pt}
    \label{tab:ablation3}
\end{table}

In the final phase of our evaluation, we scrutinized the impact of analysis on all players, teammates and adversaries. In each configuration, our agents assumed the roles of the evil side in ten games, facing off against baseline agents aided by corresponding analysis information. The results, encompassing winning rate, quest engagement rate, and failure voting rate, are tabulated in Table \ref{tab:ablation3}.
It becomes apparent that when analysis information is restricted solely to teammates, the winning rate declines by 10\%. In response, our proposed AI agents adopt a less aggressive approach, evident in reduced quest engagement rates and failure voting ratings. However, when analysis information pertains exclusively to adversaries, there is a decrease in quest engagement rates while retaining the winning rate and failure voting rate. This phenomenon can be attributed to the strategic advantage gained by the Assassin, who can identify Merlin with the aid of analysis information on adversaries. Consequently, the analysis of adversaries proves to be paramount for the evil side's victory in Avalon games for AI agents.

\section{Case Study}
\begin{figure}
    \centering
    \includegraphics[scale=0.14]{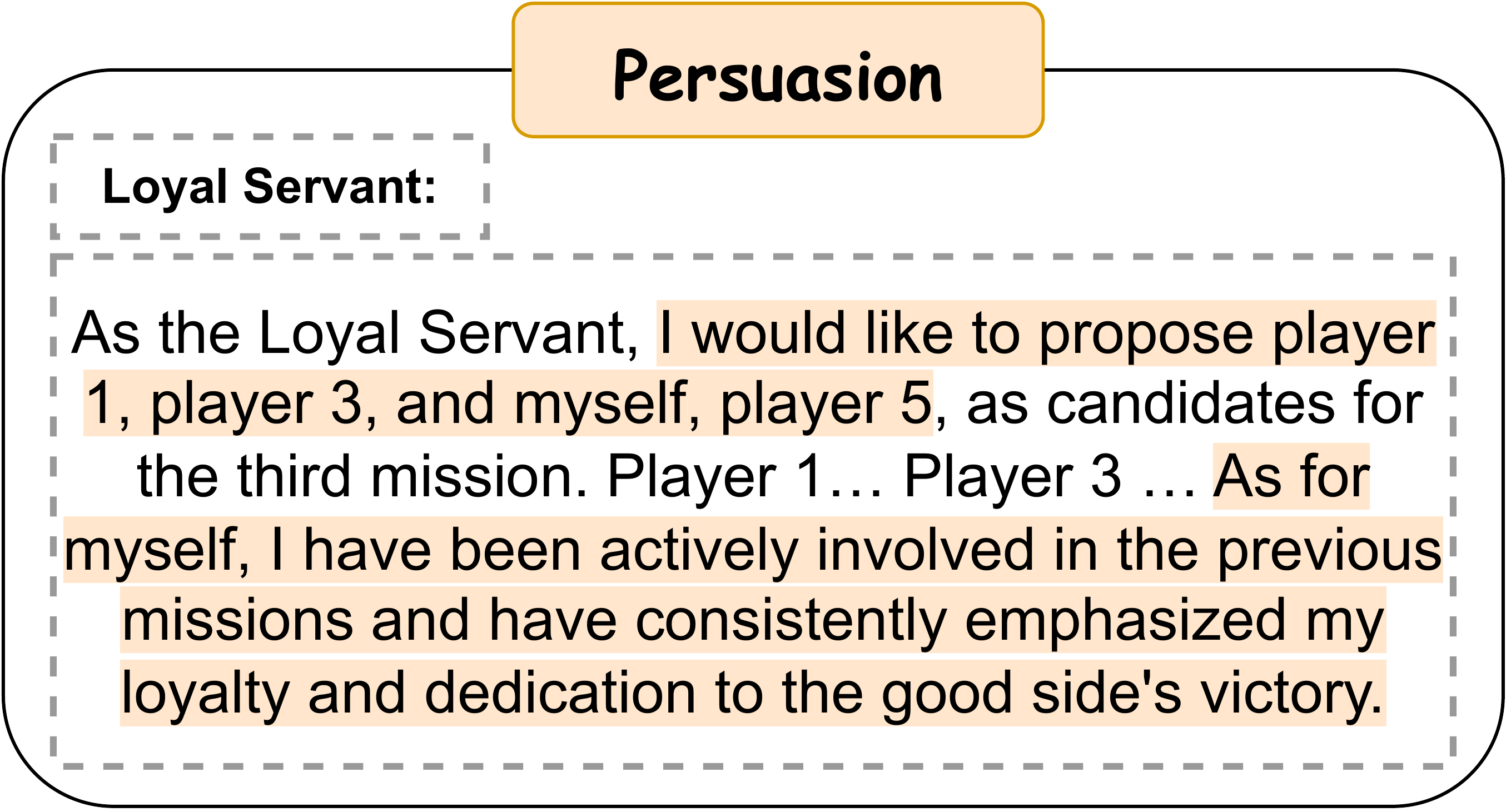}
    \vspace{-10pt}
    \caption{Persuasion example }
    \vspace{-10pt}
    \label{fig:persuasion_example}
\end{figure}

\begin{figure}
    \centering
    \includegraphics[scale=0.14]{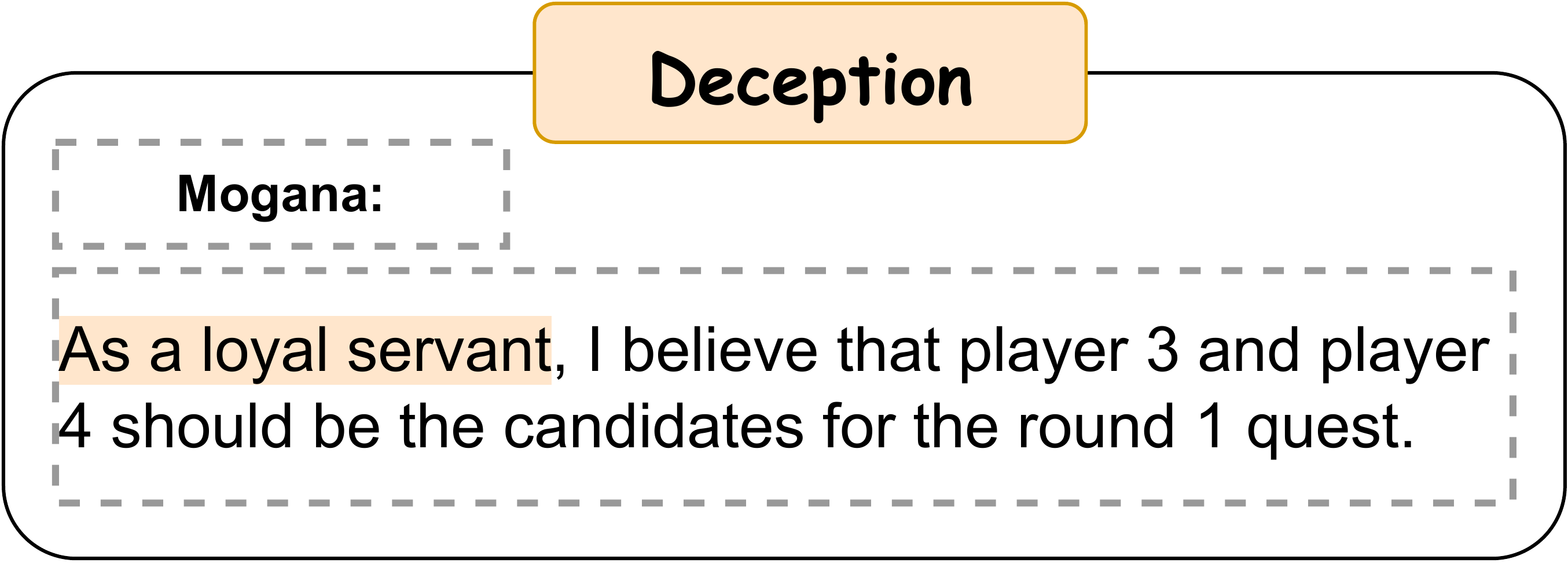}
    \vspace{-10pt}
    \caption{Camouflage example }
    \vspace{-10pt}
    \label{fig:deception_example}
\end{figure}

\begin{figure}
    \centering
    \includegraphics[scale=0.14]{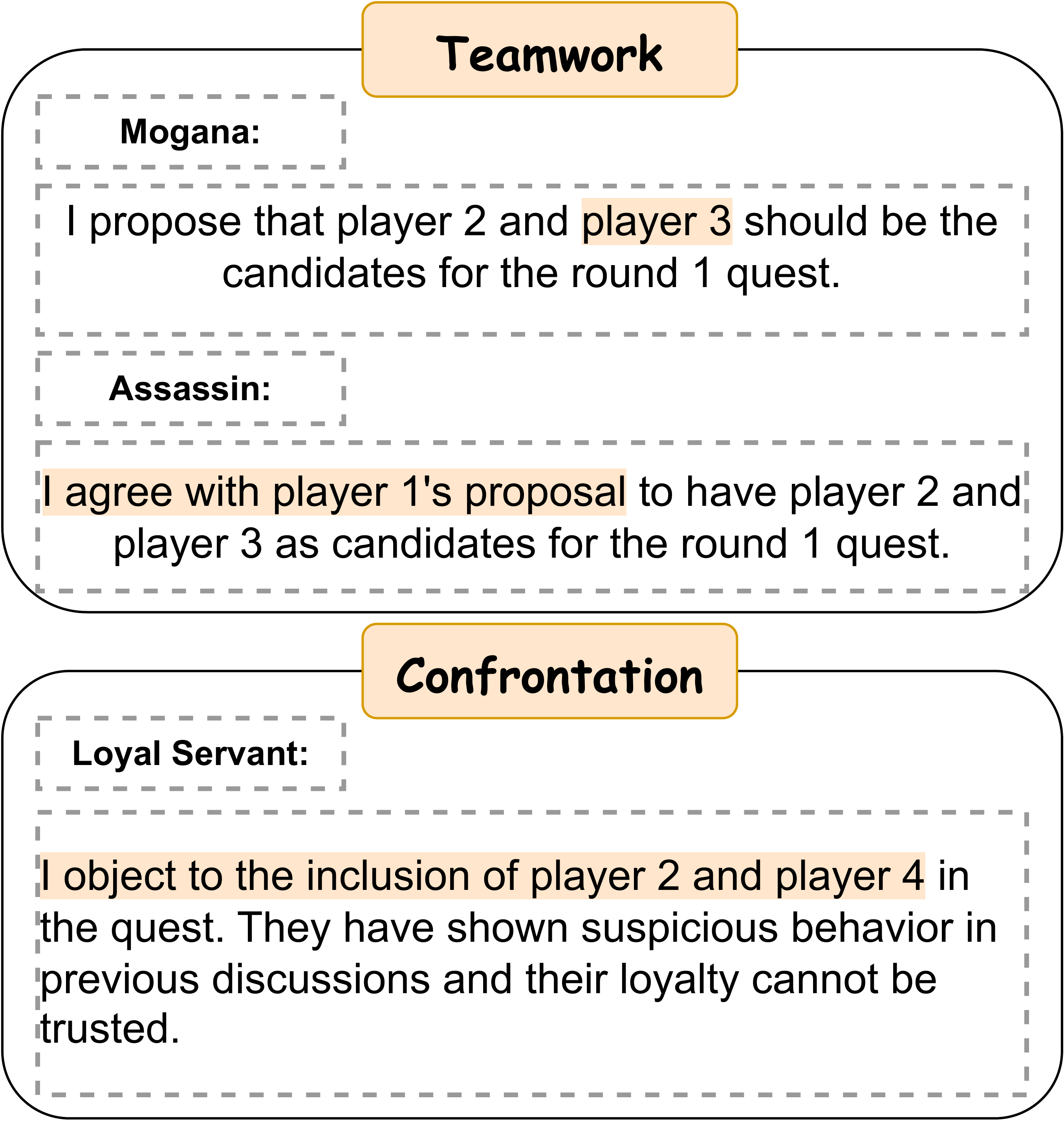}
    \vspace{-10pt}
    \caption{Teamwork and confrontation examples }
    % \vspace{-10pt}
    \label{fig:teamwork_confrontation_example}
\end{figure}

\begin{figure}
    \centering
    \includegraphics[scale=0.14]{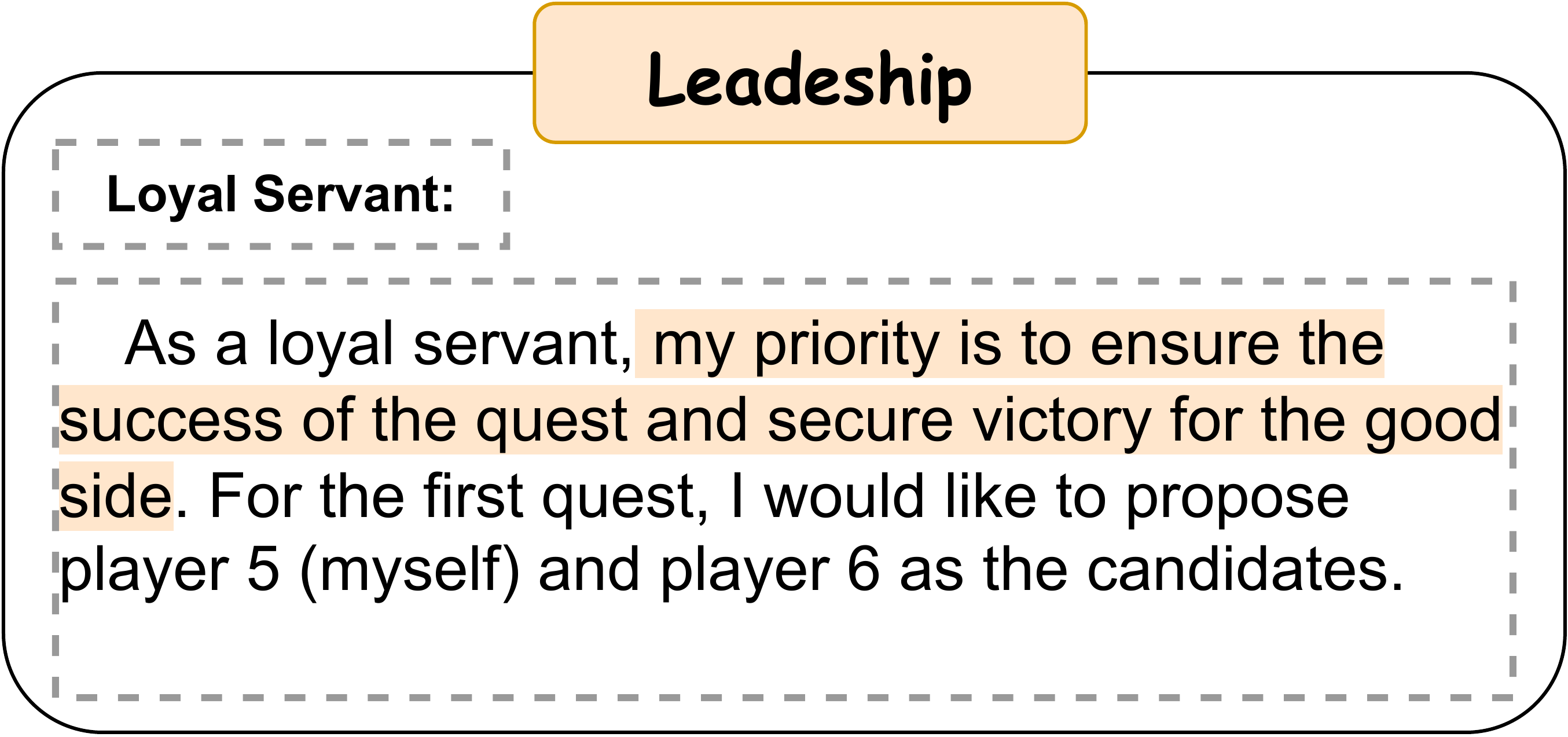}
    \vspace{-10pt}
    \caption{Leadership example }
    \vspace{-10pt}
    \label{fig:leadership_example}
\end{figure}

In Figures \ref{fig:persuasion_example}, \ref{fig:deception_example}, \ref{fig:teamwork_confrontation_example} and \ref{fig:leadership_example}, we present examples to show how the AI agents perform the social behaviors in the Avalon games. 

\section{Exploration on LLaMA-Based Agents}\label{sec:appendix_llama}

\begin{table}[]
    \centering
    \vspace{-10pt}
    \resizebox{0.46\textwidth}{!}{ 
    \begin{tabular}{c|ccccccc}
    \toprule
        Base Model & \multicolumn{6}{|c}{VRR (\%)} \\
        & Loyal Servant&Merlin&Percival&Morgana&Assassin&Average\\
        \midrule
        LLaMA2 & 51.9 & 61.0& 53.6 & 66.5& 66.9 & 59.9\\
        GPT-3.5 & 81.7 & 84.2& 81.9& 89.7& 87.6& 85.0\\
    \bottomrule
    \end{tabular}
        }
    \caption{Valid Response Rate (VRR) of different models}
    \vspace{-20pt}
    \label{tab:comparison_gpt_llama}
\end{table}

For broader validation, we implemented our framework on the Llama2-7b-chat-hf model. However, LLaMA-based agents face constraints due to the model's language understanding capabilities and token limitations. Preliminary exploration without further analysis is discussed below.

Table \ref{tab:comparison_gpt_llama} presents the performance of agents based on LLaMA2 in the Avalon game, where we measure their performance using Valid Response Rate (defined in equation \ref{eq:vrr}). Compared to GPT3.5, LLaMA shows a decrease of 25.1\% in this metric. This could be attributed to LLaMA's poorer language comprehension abilities compared to GPT3.5, resulting in its inability to grasp the complex content of the Avalon game.

\noindent\textbf{Valid Response Rate (VRR)}. Agents are required to engage in discussion, select players, and vote. A Valid Response is defined as a response that adheres to these requirements. the VRR is calculated as follows:
% The quest outcome is determined by the success or failure cards submitted by the quest team members. The failure vote rate refer the percentage of votes against the success of a quest that leads to its failure. The failure vote rate is computed with the following equation:
\begin{equation}\label{eq:vrr}
    VRR = (\frac{\# Valid\ Responses}{ \# Total\ Responses}) \times 100\%
\end{equation}

\section{Teamwork and Confrontation}\label{sec:appendix_teamwork_confrontation}
\begin{figure*}
    \centering
    \includegraphics[width=0.95\linewidth]{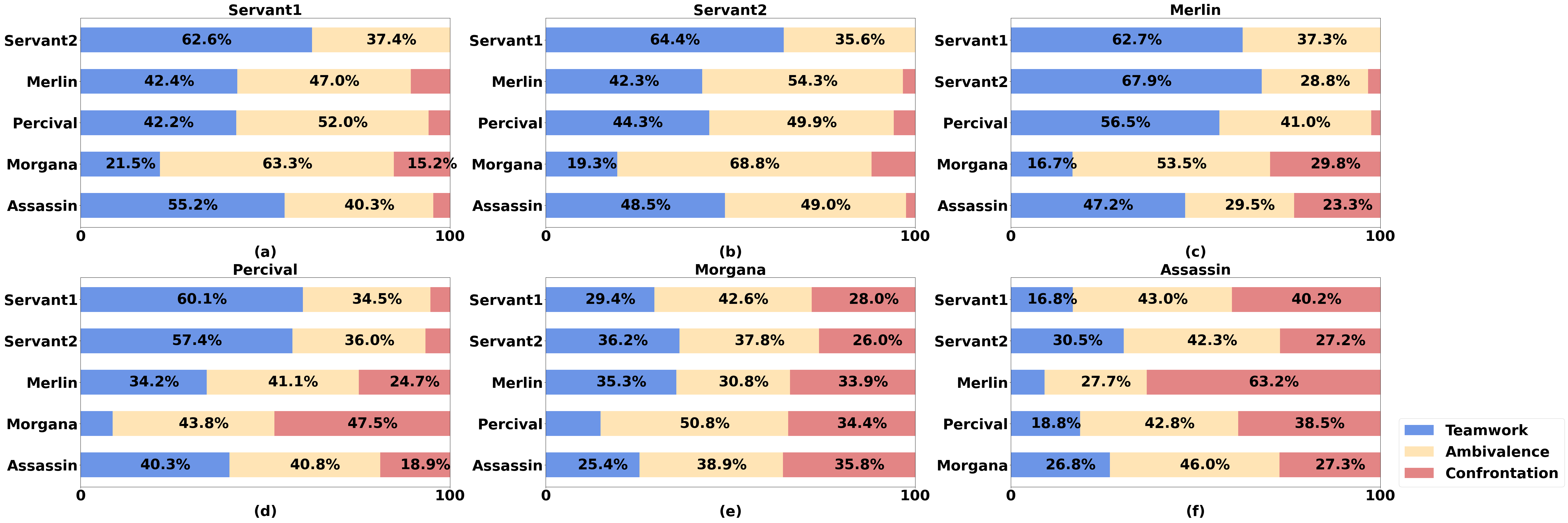}
    % \vspace{-15pt}
    \caption{The teamwork and confrontation behaviors when playing different roles: each subfigure shows the attitude distribution of the player portraying specific role (on the top) towards players in other roles (on the left). }
    % \vspace{-15pt}
    \label{fig:teamwork_confrontation_ours}
\end{figure*}

% \begin{figure*}
%     \centering
%     \includegraphics[width=0.95\linewidth]{figures/teamwork_confrontation_baseline.png}
%     \vspace{-15pt}
%     \caption{The teamwork and confrontation behaviors when playing different roles (Baseline): each subfigure shows the attitude distribution of the player portraying specific role (on the top) towards players in other roles (on the left).}
%     \vspace{-15pt}
%     \label{fig:teamwork_confrontation_baseline}
% \end{figure*}

\begin{figure*}
    \centering
    \includegraphics[width=0.95\linewidth]{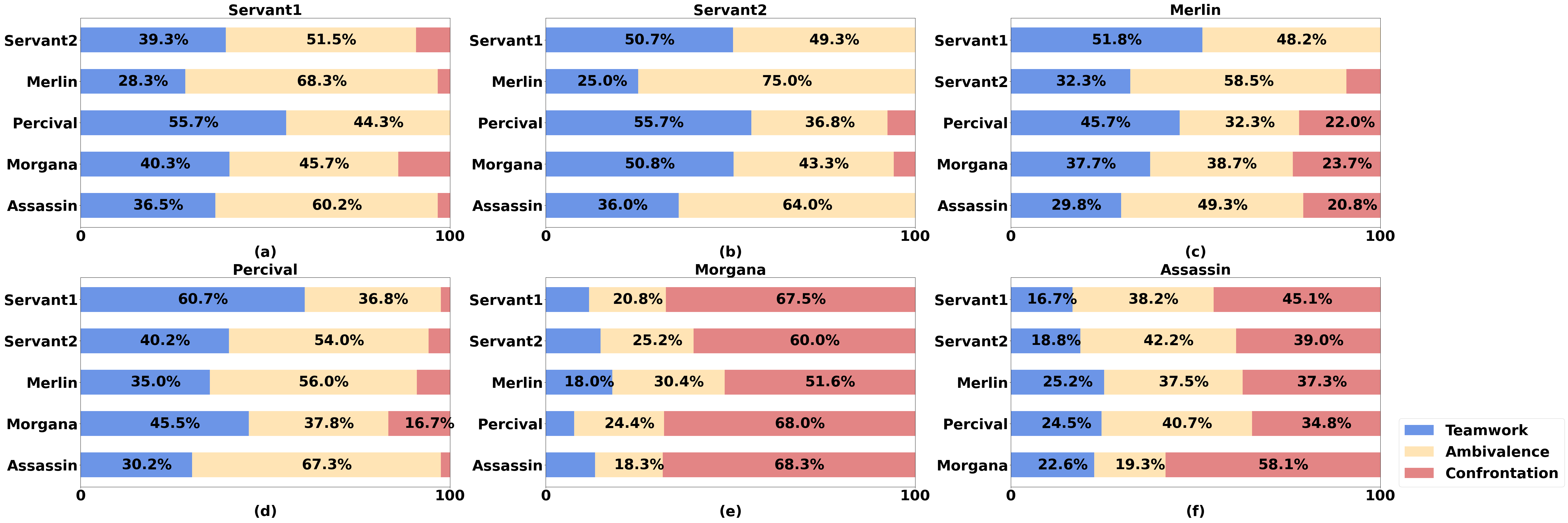}
    % \vspace{-15pt}
    \caption{The teamwork and confrontation behaviors when playing different roles (agents without experience learning module)}
    % \vspace{-15pt}
    \label{fig:teamwork_confrontation_ablation_wo_strategy}
\end{figure*}

Figure \ref{fig:teamwork_confrontation_ours} and Figure \ref{fig:teamwork_confrontation_ablation_wo_strategy} illustrate the differences in teamwork and confrontation behaviors of agents under conditions with and without experience learning.

Figure \ref{fig:teamwork_confrontation_ablation_wo_strategy} shows that, without strategic learning, evil-side players (e.g., Morgana) overly confront, while good-side players confront less, with minimal variation. This contrasts with Figure \ref{fig:teamwork_confrontation_ours}, depicting agents with strategic learning. Here, the introduction of strategic learning mitigates excessive confrontation by evil-side players, who strategically engage in more teamwork. Conversely, good-side players strategically increase confrontation with potential enemies while reducing it with potential teammates.
% \section{Example Appendix}
% \label{sec:appendix}

% This is an appendix.

\end{document}